\documentclass{article}

\usepackage[final]{neurips_2019}

\usepackage[utf8]{inputenc} 
\usepackage[T1]{fontenc}    
\usepackage{hyperref}       
\usepackage{url}            
\usepackage{booktabs}       
\usepackage{amsfonts}       
\usepackage{nicefrac}       
\usepackage{microtype}      
\usepackage{enumerate}      

\usepackage{savesym}
\savesymbol{iint} 
\savesymbol{iiint} 
\savesymbol{iiiint} 
\savesymbol{idotsint} 

\usepackage{txfonts}        
\restoresymbol{TXF}{iint}
\restoresymbol{TXF}{iiint}
\restoresymbol{TXF}{iiiint}
\restoresymbol{TXF}{idotsint}

\usepackage{xcolor}
\usepackage{graphicx}
\usepackage{mathtools}

\usepackage{algorithm}
\usepackage{algorithmic}
\usepackage{enumitem}
\usepackage{multirow}
\usepackage{caption}
\usepackage{csquotes}
\usepackage{mathabx}
\usepackage[normalem]{ulem}
\usepackage{xcolor,colortbl, makecell}
\usepackage{multicol}
\usepackage{soul}
\usepackage{framed}

\setcitestyle{authoryear,round,citesep={;},aysep={,},yysep={;}}

\newcommand{\modelname}{{\textsc{Grover}}}
\newcommand{\modelnamefordisc}{{\textsc{Grover}}}
\newcommand{\modelnamelong}{{\textbf{G}enerating a\textbf{R}ticles by \textbf{O}nly \textbf{V}iewing m\textbf{E}tadata \textbf{R}ecords}}
\newcommand{\datasetname}{{\textsc{RealNews}}}

\newcommand{\commentoutforneurips}[1]{%
}

\title{Defending Against Neural Fake News}

\author{Rowan Zellers$^\spadesuit$, Ari Holtzman$^\spadesuit$, Hannah Rashkin$^\spadesuit$, Yonatan Bisk$^\spadesuit$ \\ \textbf{Ali Farhadi$^{\spadesuit\heartsuit}$, Franziska Roesner$^\spadesuit$, Yejin Choi$^{\spadesuit\heartsuit}$}\\
  $^\spadesuit$Paul G. Allen School of Computer Science \& Engineering, University of Washington \\
  $^\heartsuit$Allen Institute for Artificial Intelligence \\
  \url{https://rowanzellers.com/grover}
}
\begin{document}

\maketitle

\begin{abstract}
 \vspace{-2mm}
Recent progress in natural language generation has raised dual-use concerns. While applications like summarization and translation are positive, the underlying technology also might enable adversaries to generate \emph{neural fake news}: targeted propaganda that closely mimics the style of real news.

Modern computer security relies on careful \emph{threat modeling}: identifying potential threats and vulnerabilities from an adversary's point of view, and exploring potential mitigations to these threats.
Likewise, developing robust defenses against neural fake news requires us first to carefully investigate and characterize the risks of these models. We thus present a model for controllable text generation called \modelname. Given a headline like `Link Found Between Vaccines and Autism,' \modelname~can generate the rest of the article; humans find these generations to be more trustworthy than human-written disinformation.

Developing robust verification techniques against generators like \modelname~is critical.
We find that best current discriminators can classify neural fake news from real, human-written, news with 73\% accuracy, assuming access to a moderate level of training data. Counterintuitively, the best defense against \modelname~turns out to be \modelname~itself, with 92\% accuracy, demonstrating the importance of public release of strong generators. We investigate these results further, showing that exposure bias -- and sampling strategies that alleviate its effects -- both leave artifacts that similar discriminators can pick up on.
We conclude by discussing ethical issues regarding the technology, and plan to release \modelname~publicly, helping pave the way for better detection of neural fake news.
 \vspace{-2mm}
\end{abstract}

\section{Introduction}
\begin{figure}[b!]
\vspace{-3mm}
  \centering\small
    \includegraphics[width=1\columnwidth]{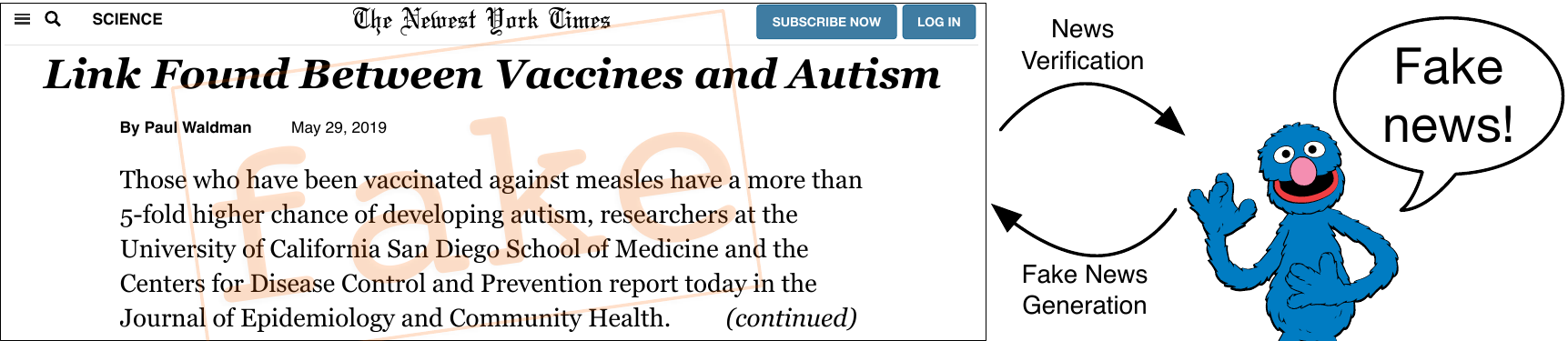}
    \vspace{-4mm}
\caption{In this paper, we explore \modelname, a model which can detect \emph{and generate} neural fake news. Humans find the articles difficult to distinguish from ``real news'' without high levels of scrutiny.}
\vspace{-5mm}
  \label{fig:teaser}
\end{figure}

Online fake news -- news designed to intentionally deceive -- has recently emerged as a major societal problem. Malicious actors spread fallacious viral stories in order to gain advertising revenue, influence opinions, and even tip elections \citep{faris2017partisanship, wardle2017information}. As such, countering the spread of disinformation online presents an urgent technical and political issue.

To the best of our knowledge, most disinformation online today is manually written \citep{vargo2018agenda}. However, as progress continues in natural language generation, malicious actors will increasingly be able to controllably generate realistic-looking propaganda at scale. Thus, while we are excited about recent progress in text generation \citep{Jzefowicz2016ExploringTL,radford2018improving,radford2019gpttwo}, we are also concerned with the inevitability of AI-generated `neural' fake news.\footnote{
We thank past work, such as \href{https://openai.com/blog/better-language-models/}{OpenAI's Staged Release Policy for GPT2} for drawing attention to neural disinformation, alongside other dual-use implications.
}

With this paper, we seek to understand and respond to neural fake news \emph{before} it manifests at scale. We draw on the field of computer security, which relies on \emph{threat modeling}: analyzing the space of potential threats and vulnerabilities in a system to develop robust defenses. 
To scientifically study the risks of neural disinformation, we present a new generative model called \modelname.\footnote{Short for \modelnamelong.} Our model allows for controllable yet efficient generation of an entire news article -- not just the body, but also the title, news source, publication date, and author list. This lets us study an adversary with controllable generations 
(e.g. Figure~\ref{fig:teaser}, an example anti-vaccine article written in the style of the New York Times).

Humans rate the disinformation generated by \modelname~as trustworthy, even more so 
than human-written disinformation. Thus, developing robust verification techniques against generators such as \modelname~is an important research area.
We consider a setting in which a discriminator has access to 5000 \modelname~generations, but unlimited access to real news. In this setting, the best existing fake news discriminators 
are, themselves, deep pretrained language models (73\% accuracy) \citep{peters2018deep,radford2018improving,radford2019gpttwo,devlin2018bert}. However, we find that \modelname,~when used in a discriminative setting, performs even better at 92\% accuracy. This finding represents an exciting opportunity for defense against neural fake news: the best models for generating neural disinformation are also the best models at detecting it. 

Next, we investigate how deep pretrained language models distinguish between real and machine-generated text. We find that key artifacts are introduced during generation as a result of exposure bias: the generator is not perfect, so randomly sampling from its distribution results in generations that fall increasingly out-of-distribution as length increases. However, sampling strategies that alleviate these effects also introduce artifacts that strong discriminators can pick up on.

We conclude with a sketch of the ethical territory that must be mapped out in order to understand our responsibilities as researchers when studying fake news, and the potential negative implications of releasing models \citep{hecht2018s, zellers2019whywereleasedgrover, solaiman2019release}.
Accordingly, we suggest a provisional policy of how such models should be released and why we believe it to be safe  -- and perhaps even imperative -- to do so.  We believe our proposed framework and accompanying models provide a concrete initial proposal for an evolving conversation about ML-based disinformation threats and how they can be countered.

\section{Fake News in a Neural and Adversarial Setting}
\label{sec:overview}
We present a framework -- motivated by today's dynamics of manually created fake news -- for understanding what \emph{adversaries} will attempt with deep models, and how \emph{verifiers} should respond.

\paragraph{Scope of fake news.}
There are many types of \emph{false} news, ranging from satire to propaganda \citep{wardle2017fake}. In this paper, we focus on text-only documents formatted as news articles: stories and their corresponding metadata that contain purposefully false information.
Existing fake news is predominantly human-written, for two broad goals: monetization (ad revenue through clicks) and propaganda (communicating targeted information)  \citep{bradshaw2017troops, melford2019disinfo}. Achieving either goal requires the adversary to be selective about the news that they make, whether by producing only viral content, or content that advances a given agenda.

\paragraph{Fact checking and verification: related work.}
There is considerable interest in fighting online disinformation. Major platforms such as Facebook prioritize trustworthy sources and shut down accounts linked to disinformation \citep{mosseri2018news,dwoskin2018facebook}. Some users of these platforms avoid fake news with tools such as NewsGuard and Hoaxy \citep{shao2016hoaxy} and websites like Snopes and PolitiFact. These services rely on manual fact-checking efforts: verifying the accuracy of claims, articles, and entire websites. 
Efforts to automate fake news detection generally point out stylistic biases that exist in the text \citep{rashkin2017truth, wang2017liar,perez2018automatic}. These efforts can help moderators on social media platforms shut down suspicious accounts. However, fact checking is not a panacea -- cognitive biases such as the backfire effect and confirmation bias make humans liable to believe fake news that fits their worldview \citep{swire2017role}.

\paragraph{Framework.}\hspace{-1mm} We cast fake news generation and detection as an adversarial game, with two players:

\begin{itemize}[wide, leftmargin=10pt, labelwidth=!,labelindent=0pt,noitemsep,topsep=0pt]
    \item \textbf{Adversary}. Their goal is to generate fake stories that match specified attributes: generally, being viral or persuasive. The stories must read realistically to both human users as well as the verifier. 
    \item \textbf{Verifier}. Their goal is to classify news stories as real or fake. The verifier has access to unlimited real news stories, but few fake news stories from a specific adversary. This setup matches the existing landscape: when a platform blocks an account or website,
    their disinformative stories provide training for the verifier; but 
    it is difficult to collect fake news from newly-created accounts.
\end{itemize}

The dual objectives of these two players suggest an escalating ``arms race'' between attackers and defenders. As verification systems get better, so too will adversaries. We must therefore be prepared to deal with ever-stronger adversarial attacks, which is the focus of the next section.

\section{\modelname: Modeling Conditional Generation of Neural Fake News}
\definecolor{domain}{HTML}{FFC7BF}
\definecolor{date}{HTML}{FFE9BF}
\definecolor{authors}{HTML}{CFFFCC}
\definecolor{headline}{HTML}{C0DEFF}
\definecolor{body}{HTML}{E3C0FF}
\definecolor{boringcolor}{rgb}{0.95, 0.95, 0.95}

\makeatletter
 \def\SOUL@hlpreamble{%
 \setul{}{2.4ex}
 \let\SOUL@stcolor\SOUL@hlcolor
 \SOUL@stpreamble
 }
\makeatother

\newcommand{\hlc}[2][yellow]{{%
    \colorlet{foo}{#1}%
    \sethlcolor{foo}\hl{#2}}%
}

\newcommand{\metadata}{metadata}
\newcommand{\Metadata}{Metadata}

\newcommand{\tokenstart}{{\tt\small <start>}}
\newcommand{\tokenend}{{\tt\small <end>}}
\newcommand{\taustart}{{\tt\small <start${-}\tau$>}}
\newcommand{\tauend}{{\tt\small <end${-}\tau$>}}

\newcommand{\bodyfield}{\hlc[body]{body}}
\newcommand{\domainfield}{\hlc[domain]{domain}}
\newcommand{\datefield}{\hlc[date]{date}}
\newcommand{\authorsfield}{\hlc[authors]{authors}}
\newcommand{\authorfield}{\hlc[authors]{author}}
\newcommand{\headlinefield}{\hlc[headline]{headline}}
Given existing 
online disinformation, we have reason to believe adversaries will try to generate targeted content (e.g. clickbait and propaganda). Recently introduced large-scale generative models produce realistic-looking text \citep{radford2019gpttwo}, but they do not lend themselves to producing controllable generations \citep{hu2017toward}.\footnote{A common workaround is to have a human seed the text to provide context. However, this \textbf{a)} is a heavy handed technique for biasing which may not capture the desired attributes, and \textbf{b)} leaves in place a human-written beginning (as tokens are only generated left-to-right), which may create distributional artifacts.} 
Therefore, to probe the feasibility of realistic-looking neural fake news, we introduce \modelname, which produces both realistic \emph{and} controlled generations.

The current state-of-the-art in unconditional text generation views it as a language modeling problem \citep{bengio2003neural}, in which the probability of a document $\boldsymbol{x}$ is the product of the conditional probability of generating each token $x_{i}$ given previous tokens:
{
\setlength{\abovedisplayskip}{1pt}
\setlength{\belowdisplayskip}{1pt}
\setlength{\abovedisplayshortskip}{0pt}
\setlength{\belowdisplayshortskip}{0pt}
\begin{equation}
\label{eq:lmodeling}
    p(\boldsymbol{x}) = \prod_{i=1}^{N} p(x_i | x_1 \ldots x_{i-1}).
\end{equation}
}%
The document is typically treated as a single unstructured \emph{text field}, beginning with a \tokenstart~token and ending with an \tokenend~token. The latter, \tokenend, is particularly important because it indicates the end of the field, and when to should stop generating.
However, a news article has necessary structure beyond the running text, or \bodyfield~field. \Metadata~fields include the \domainfield~where the article is published (indirectly marking the style), the \datefield~of publication, the names of the \authorsfield, and the \headlinefield~of the article itself. Not only does generating a news article require producing all of these components, these fields also allow significant control over the generations (e.g. specifying a \headlinefield~helps control the generated \bodyfield). An article can be modeled by the joint distribution:
{
\setlength{\abovedisplayshortskip}{0pt}
\setlength{\belowdisplayshortskip}{0pt}
\begin{equation}
\label{eq:jointdist}
p(\textrm{\domainfield}, \textrm{\datefield}, \textrm{\authorsfield}, \textrm{\headlinefield}, \textrm{\bodyfield}).
\end{equation}
}%
However, it is not immediately obvious how to sample from Equation~\ref{eq:jointdist}. One option is to define a \emph{canonical order} among the article's fields $\mathcal{F}$: ($f_1{\prec}f_2{\prec}{\ldots}{\prec}f_{|\mathcal{F}|}$), and model the article left-to-right in that order using Equation~\ref{eq:lmodeling}: $x^{f_1}_{1}, x^{f_1}_{2}, \ldots, x^{f_{|\mathcal{F}|}}_{|f_{|\mathcal{F}|}|}$. However, this ordering would forbid sampling certain fields without prohibitively expensive marginalization. Alternatively, one could generate fields in any order, but this requires the model to learn to handle $|\mathcal{F}|!$ potential orderings during inference time. 

Our solution is \modelname, a new approach for efficient learning and generation of multi-field documents. We adopt the language modeling framework of Equation~\ref{eq:lmodeling} in a way that allows for flexible decomposition of Equation~\ref{eq:jointdist}. During inference time, we start with a set of fields $\mathcal{F}$ as context, with each field $f$ containing field-specific start and end tokens. We sort the fields using a standard order\footnote{Our ordering is the following field types in order:  \domainfield, \datefield, \authorsfield, \headlinefield, and then the \bodyfield.} and combine the resulting tokens together. To generate a target field $\tau$, we append the field-specific start token \taustart~to the context tokens; then, we sample from the model until we hit \tauend.

\begin{figure}[t!]
  \centering\small
    \includegraphics[width=\textwidth]{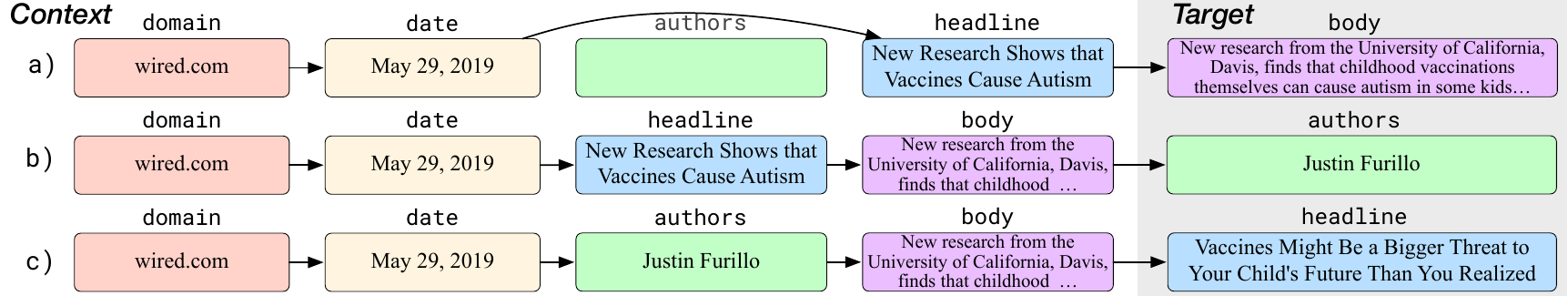}
    \vspace{-4mm}
\caption{A diagram of three \modelname~examples for article generation. In row {\tt\small a)}, the \hlc[body]{body} is generated from partial context (the \hlc[authors]{authors} field is missing). In {\tt\small b)}, the model generates the \hlc[authors]{authors}. In {\tt\small c)}, the model uses the new generations to regenerate the provided \hlc[headline]{headline} to one that is more realistic.
}\vspace{-3mm}
  \label{fig:setup}
\end{figure}

Figure \ref{fig:setup} shows an example of using \modelname~to generate an anti-vaccine article. Here, the adversary specifies a \domainfield, \datefield, and \headlinefield. After \modelname~generates the \bodyfield, it can be used to generate a fake \authorfield, before finally generating a new and more appropriate \headlinefield.

During training, we simulate inference by randomly partitioning an article's fields into two disjoint sets $\mathcal{F}_1$ and $\mathcal{F}_2$. We also randomly drop out individual fields with probability 10\%, and drop out all but the \bodyfield~with probability 35\%. This allows the model to learn how to perform unconditional generation. We sort the metadata fields in each set using our standard order, and concatenate the underlying tokens. The model is then trained to minimize the cross-entropy of predicting the tokens in $\mathcal{F}_1$ followed by the tokens in $\mathcal{F}_2$.\footnote{All tokens use the same vocabulary. By using a standard order, but partitioning the fields into two sets, the model can generate any field conditioned on others while only needing to learn $2^{|\mathcal{F}|}$ orderings, versus $|\mathcal{F}|!$.}

\paragraph{Architecture.}
We draw on recent progress in training large Transformers for language modeling \citep{vaswani2017attention}, building \modelname~using the same architecture as for GPT2 \citep{radford2019gpttwo}. We consider three model sizes. Our smallest model, \modelname-Base, has 12 layers and 124 million parameters, on par with GPT and BERT-Base \citep{radford2018improving,devlin2018bert}. Our next model, \modelname-Large, has 24 layers and 355 million parameters, on par with BERT-Large. Our largest model, \modelname-Mega, has 48 layers and 1.5 billion parameters, on par with GPT2.

\paragraph{Dataset.} We present \datasetname, a large corpus of news articles from Common Crawl. Training \modelname~requires a large corpus of news articles with metadata, but none currently exists. Thus, we construct one by scraping dumps from Common Crawl, limiting ourselves to the ~5000 news domains indexed by Google News. We used the Newspaper Python library to extract the \bodyfield~and~\metadata~from each article. News from Common Crawl dumps from December 2016 through March 2019 were used as training data; articles published in April 2019 from the April 2019 dump were used for evaluation. After deduplication, \datasetname~is 120 gigabytes without compression.

\paragraph{Learning.} We trained each \modelname~model on randomly-sampled sequences from \datasetname~with length 1024. Other optimization hyperparameters are in Appendix~\ref{sec:optimizationhyperparameters}. We trained \modelname-Mega for 800k iterations, using a batch size of 512 and 256 TPU v3 cores.
Training time was two weeks.

\subsection{Language Modeling results: measuring the importance of data, context, and size}
We validate \modelname, versus standard unconditional language models, on the  April 2019 test set. We consider two evaluation modes: \emph{unconditional}, where no context is provided and the model must generate the article \bodyfield; and \emph{conditional}, in which the full metadata is provided as context. In both cases, we calculate the perplexity only over the article \bodyfield.

Our results, shown in Figure~\ref{fig:ppl}, show several conclusions. First, \modelname~noticeably improves (between .6 to .9 perplexity points) when conditioned on metadata. Second, perplexity decreases with size, with \modelname-Mega obtaining 8.7 perplexity in the conditional setting. Third, the data distribution is still important: though the GPT2 models with 124M parameters and 355M parameters respectively match our \modelname-Base and \modelname-Large architectures, our model is over 5 perplexity points lower in both cases, possibly because the OpenAI WebText corpus also contains non-news articles.

\subsection{Carefully restricting the variance of generations with Nucleus Sampling}
Sampling from \modelname~is straightforward as it behaves like a left-to-right language model during decoding. However, the choice of decoding algorithm is important.
While likelihood-maximization strategies such as beam search work well for \emph{closed-ended} generation tasks where the output contains the same information as the context (like machine translation), these approaches have been shown to produce degenerate text during \emph{open-ended} generation \citep{hashimoto2019unifying, holtzman2019curious}. However, as we will show in Section~\ref{sec:analysis}, restricting the variance of generations is also crucial. 
%

In this paper, we primarily use Nucleus Sampling (top-$p$): for a given threshold $p$, at each timestep we sample from the most probable words whose cumulative probability comprises the top-$p$\% of the entire vocabulary  \citep{holtzman2019curious}.\footnote{In early experiments, we found Nucleus Sampling produced better and less-detectable generations than alternatives like top-$k$ sampling, wherein the most probable $k$ tokens are used at each timestep \citep{fan2018hierarchical}.}

\section{Humans are Easily Fooled by \modelname-written Propaganda}
\label{sec:genexps}
\begin{figure}[t!]
\centering\small
\begin{minipage}{.48\textwidth}
  \centering\small
  \includegraphics[width=\linewidth]{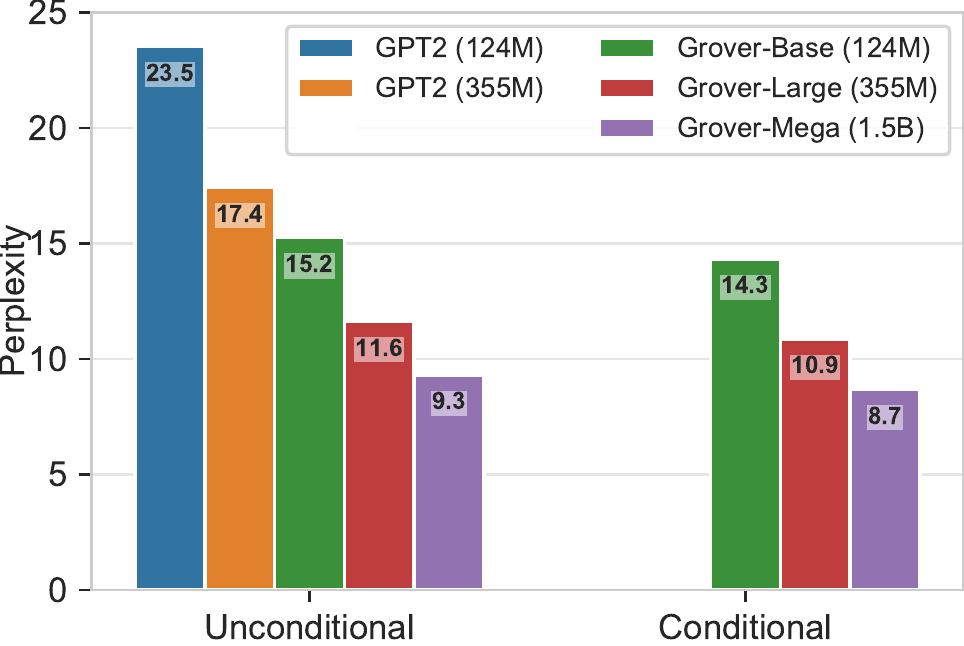}\vspace*{-1mm}
  \captionof{figure}{Language Modeling results on the \bodyfield~field of April 2019 articles. We evaluate in the \emph{Unconditional} setting (without provided metadata) as well as in the \emph{Conditional} setting (with all metadata). \modelname~sees over a 0.6 point drop in perplexity when given metadata.}
  \label{fig:ppl}
\end{minipage}%
\hspace{.039\textwidth}\begin{minipage}{.48\textwidth}\vspace*{-3mm}
  \centering\small
  \includegraphics[width=\linewidth]{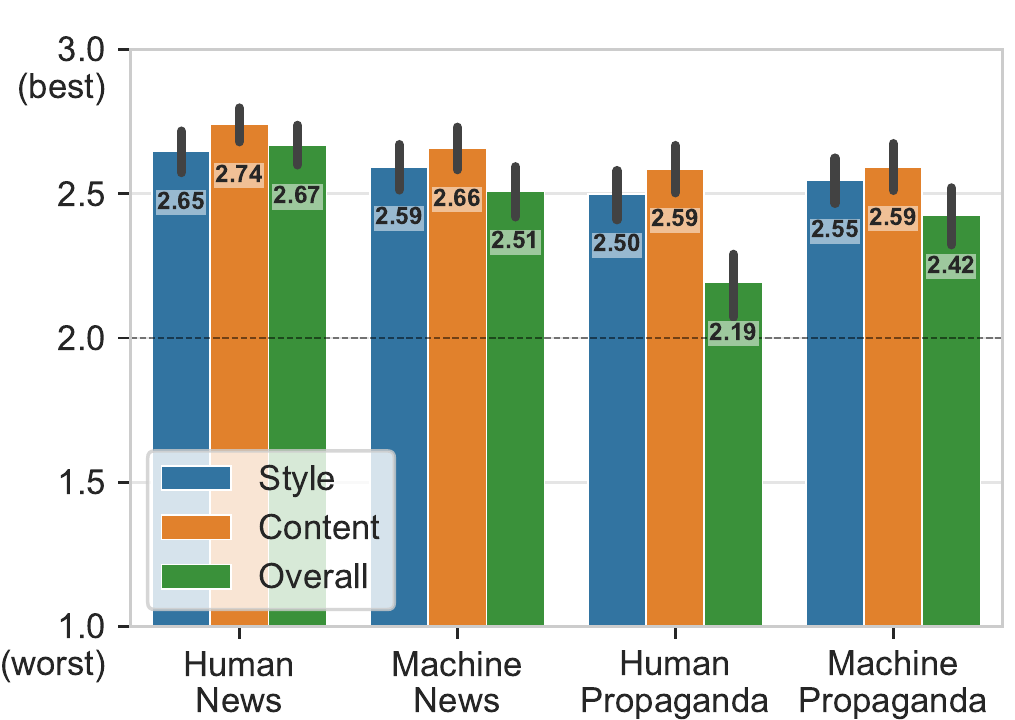}\vspace*{-2mm}
  \captionof{figure}{Human evaluation. For each article, three annotators evaluated style, content, and the overall trustworthiness; 100 articles of each category were used. The results show that propaganda generated by \modelname~is rated more plausible than the original human-written propaganda.}\vspace{-1mm}
  \label{fig:humaneval}
\end{minipage}\vspace{-4mm}\end{figure}

We evaluate the quality of disinformation generated by our largest model, \modelname-Mega, using $p{=}.96$. We consider four classes of articles: 
human-written articles from reputable news websites ({\tt\small Human News}), 
\modelname-written articles conditioned on the same metadata ({\tt\small Machine News}),
human-written articles from known \emph{propaganda} websites ({\tt\small Human Propaganda}),
and \modelname-written articles conditioned on the propaganda metadata ({\tt\small Machine Propaganda}).\footnote{We use the technique described in Figure~\ref{fig:setup} to rewrite the propaganda: given the metadata, generate the article first, and then rewrite the headline.} The domains used are in Appendix~\ref{sec:newssites}; examples are in Appendix~\ref{sec:suppexamples}. We asked a pool of qualified workers on Amazon Mechanical Turk to rate each article on three dimensions: stylistic consistency, content sensibility, and overall trustworthiness.\footnote{With these guidelines, we tried to separate style versus content. Overall trustworthiness asks `Does the article read like it comes from a trustworthy source?' which emphasizes style, while content sensibility asks whether the content is believable on a semantic level.}

Results (Figure~\ref{fig:humaneval}) show a striking trend: though the quality of \modelname-written news is not as high as human-written news, it is adept at rewriting propaganda. The overall trustworthiness score of propaganda increases from 2.19 to 2.42 (out of 3) when rewritten by \modelname.\footnote{This difference is statistically significant at $p=0.01$. One possible hypothesis for this effect is that \modelname~ignores the provided context. To test this hypothesis, we did a human evaluation of the consistency of the article body with the headline, date, and author. We found that human-written propaganda articles are consistent with the headline with an average score of 2.85 of 3 on the same 1-3 scale, while machine-written propaganda is consistent with 2.64 of 3.}

\section{Neural Fake News Detection}
\label{sec:detection}
The high quality of neural fake news written by \modelname, as judged by humans, makes automatic neural fake news detection an important research area. Using models (below) for the role of the \emph{Verifier} can mitigate the harm of neural fake news by classifying articles as {\tt\small Human} or {\tt\small Machine} written. These decisions can assist content moderators and end users in identifying likely (neural) disinformation. 




\begin{enumerate}[wide, leftmargin=10pt, labelwidth=!,labelindent=-2pt,itemsep=1pt,topsep=0pt,label=\textbf{\alph*}.]
    \item \modelnamefordisc. We consider a version of our model adapted for discrimination. Similar to GPT \citep{radford2018improving}, we place a special {\small\tt [CLS]} token at the end of each article, and extract the final hidden state at that point. The hidden state is fed to a linear layer to predict the label {\tt\small Human} or {\tt\small Machine}. 
    
    To simulate real conditions, and ensure minimal overlap between the generator and discriminator parameters, we initialize \modelnamefordisc~for discrimination using the checkpoint at iteration 700k, whereas the generator uses the checkpoint at iteration 800k.
    \item GPT2, a 124M or 355M parameter pretrained Transformer language model. Similar to \modelnamefordisc, we follow the GPT approach and extract the hidden state from a newly-added  {\small\tt [CLS]} token.
    \item BERT, a 110M parameter (BERT-Base) or 340M parameter (BERT-Large) bidirectional Transformer encoder commonly used for discriminative tasks. We perform domain adaptation to adapt BERT to the news domain, as well as to account for long articles; details in Appendix~\ref{sec:bertda}.
    \item FastText, an off-the-shelf library for bag-of-ngram text classification \citep{joulin2017bag}. Though not pretrained, similar models do well at detecting human-written fake news.
\end{enumerate}
All models are trained to minimize the cross-entropy loss of predicting the right label. Hyperparameters used during discrimination are in Appendix~\ref{sec:dischyperparams}.

\subsection{A semi-supervised setting for neural fake news detection}
While there are many human-written articles online, most are from the distant past, whereas articles to be detected will likely be set in the present. Likewise, there might be relatively few neural fake news articles from a given adversary.\footnote{Moreover, since disinformation can be shared on a heterogeneous mix of platforms, it might be challenging to pin down a single generated model.} We thus frame neural fake news detection as a semi-supervised problem. A neural verifier (or \emph{discriminator}) has access to many human-written news articles from March 2019 and before -- the entire \datasetname~training set. However, it has limited access to generations, and more recent news articles. Using 10k news articles from April 2019, we generate article body text; another 10k articles are used as a set of human-written news articles. We split the articles in a balanced way, with 10k for training (5k per label), 2k for validation, and 8k for testing.

We consider two evaluation modes. In the \textbf{unpaired} setting, a discriminator is provided single news articles, and must classify each independently as {\tt\small Human} or {\tt\small Machine}. In the \textbf{paired} setting, a model is given two news articles with the same metadata, one real and one machine-generated. The discriminator must assign the machine-written article a higher {\tt\small Machine} probability than the human-written article. We evaluate both modes in terms of accuracy.

\subsection{Discrimination results: \modelname~performs best at detecting \modelname's fake news}
\definecolor{lightgray}{rgb}{0.95, 0.95, 0.95}
\newcolumntype{g}{>{\columncolor{lightgray}}c}

\newcommand{\best}[1]{\textbf{#1}}
\newcommand{\scnd}[1]{#1}
\newcommand{\resultswidth}{1.35cm}

\begin{figure}[t!]
\centering\small
\begin{minipage}{.48\textwidth}
  \centering\scriptsize
    \captionof{table}{Results of discriminators versus generators, in both the paired and unpaired settings and across architecture sizes. We also vary the generation hyperparameters for each generator-discriminator pair, reporting the discrimination test accuracy for the hyperparameters with the \emph{lowest} validation accuracy. Compared with other models such as BERT, \modelname~is the best at detecting its own generations as neural fake news.}
\begin{tabular}{@{}l@{\hspace{0.25em}}r@{\hspace{0.5em}}l@{\hspace{0.25em}}c@{\hspace{0.5em}}c@{\hspace{-2em}}c@{\hspace{-2em}}||@{\hspace{0.4em}}c@{\hspace{0.25em}}c@{\hspace{0.3em}}c@{}}
& & & \multicolumn{3}{g}{Unpaired Accuracy} & \multicolumn{3}{g}{Paired Accuracy} \\
& & & \multicolumn{3}{c}{Generator size} & \multicolumn{3}{c}{Generator size} \\
& & & 1.5B & 355M & 124M & 1.5B & 355M & 124M \\
\cmidrule{3-9}
& & Chance & & 50.0 & & & 50.0 & \\ \cmidrule{3-9}
\multirow{8}{*}{\rotatebox[origin=c]{90}{Discriminator size}} & 1.5B & \modelname-Mega&\textbf{91.6}&\textbf{98.7}&\textbf{99.8}&\textbf{98.8}&\textbf{100.0}&\textbf{100.0}\\ \cmidrule{3-9}
& \multirow{3}{*}{355M} & \modelname-Large&\textbf{79.5}&\textbf{91.0}&\textbf{98.7}&\textbf{88.7}&\textbf{98.4}&\textbf{99.9}\\
& & BERT-Large&68.0&78.9&93.7&75.3&90.4&99.5\\
& & GPT2&70.1&77.2&88.0&79.1&86.8&95.0\\ \cmidrule{3-9}
& \multirow{3}{*}{124M} & \modelname-Base&\textbf{71.3}&\textbf{79.4}&\textbf{90.0}&80.8&88.5&\textbf{97.0}\\
& & BERT-Base&67.2&75.0&82.0&\textbf{84.7}&\textbf{90.9}&96.6\\
& & GPT2&67.7&73.2&81.8&72.9&80.6&87.1\\ \cmidrule{3-9}
& 11M & FastText&63.8&65.4&70.0&73.0&73.0&79.0 \\ \cmidrule{3-9}
\end{tabular}
  \label{tab:results}
\end{minipage}%
\hspace{.039\textwidth}\begin{minipage}{.48\textwidth}
  \centering\small
  \includegraphics[width=.97\linewidth]{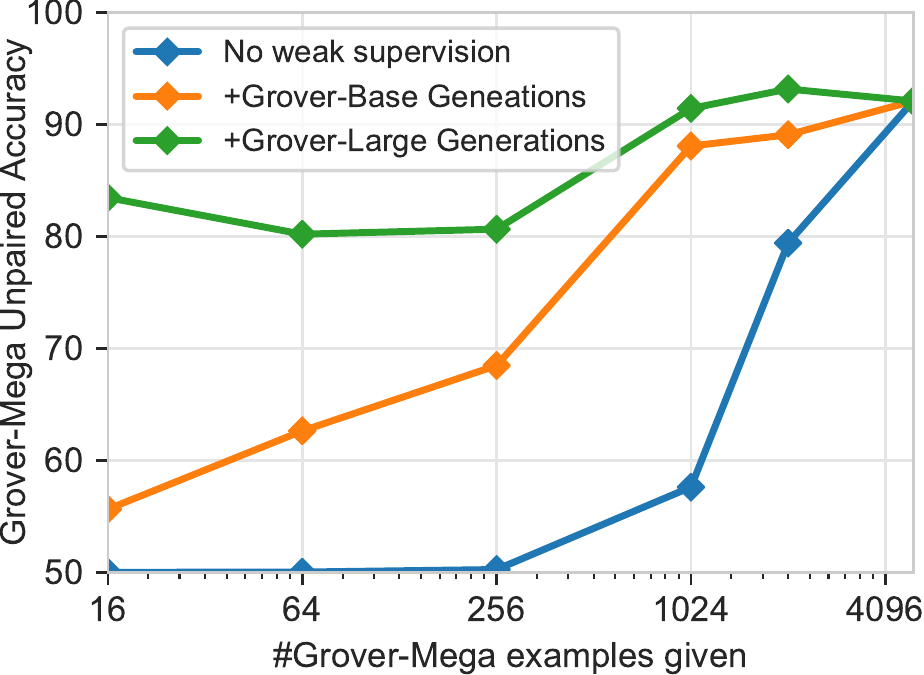} \vspace{-1mm}
  \captionof{figure}{Exploring weak supervision for discriminating \modelname-Mega generations. 
  With no weak supervision, the discriminator sees 
  $x$ 
  machine-written articles (from \modelname~Mega). For $+$\modelname-Base and $+$\modelname-Mega, the discriminator sees $5000{-}x$ machine-written articles given by the weaker generator in question. Seeing weaker generations improves performance when few in-domain samples are given. }
  \label{fig:weaksupervision}
\end{minipage}\vspace{-5mm}
\end{figure}
We present experimental results in Table~\ref{tab:results} for all generator and discriminator combinations. For each pair, we show the test results using the most adversarial generation hyperparameters (top-$p$) as judged on the validation set.\footnote{For each discriminator/generator pair, we search over $p \in \{.9,.92,.94,.96,.98,1.0\}$.} The results show several trends. First, the paired setting appears much easier than the unpaired setting, suggesting that it is difficult for the model to calibrate its predictions. Second, model size is highly important in the arms race between generators and discriminators. Using \modelname~to discriminate \modelname's generations results in roughly 90\% accuracy across the range of sizes. If a larger generator is used, accuracy slips below 81\%; conversely, if the discriminator is larger, accuracy is above 98\%. Third, other discriminators perform worse than \modelname~overall, even when controlling for architecture size and (for both BERT models) the domain. 

That \modelnamefordisc~is the best discriminator is possibly surprising: being unidirectional, it is less expressive than deep bidirectional models such as BERT.\footnote{Indeed, bidirectional approaches perform best on leaderboards like GLUE \citep{wang2018glue}.} That the more expressive model here is \textbf{not} the best at discriminating between real and generated news articles suggests that neural fake news discrimination requires having a similar \emph{inductive bias} as the generator.\footnote{This matches findings on the HellaSwag dataset \citep{zellers2018hellaswag}. Given human text and machine text written by a finetuned GPT model, a GPT discriminator outperforms BERT-Base at picking out human text.}

\subsection{Weak supervision: what happens if we don't have access to \modelname-Mega?}
These results suggest that \modelnamefordisc~is an effective discriminator when we have a medium number of fake news examples from the exact adversary that we will encounter at test time. What happens if we relax this assumption? Here, we consider the problem of detecting an adversary who is generating news with \modelname-Mega and an unknown top-$p$ threshold.\footnote{The top-$p$ threshold used was $p{=}0.96$, but we are not supposed to know this!} In this setup, during training, we have access to a weaker model (\modelname-Base or \modelname-Large). We consider the effect of having only $x$ examples from \modelname-Mega, and sampling the missing $5000{-}x$ articles from one of the weaker models, where the top-p threshold is uniformly chosen for each article in the range of $[0.9, 1.0]$. 

We show the results of this experiment in Figure~\ref{fig:weaksupervision}. The results suggest that observing additional generations greatly helps discrimination performance when few examples of \modelname-Mega are available: weak supervision with between 16 and 256 examples from \modelname-Large yields around 78\% accuracy, while accuracy remains around 50\% without weak supervision. As the portion of examples that come from \modelname-Mega increases, however, accuracy converges to around 92\%.\footnote{In additional experiments we show that accuracy increases even more -- up to 98\% -- when the number of examples is increased \citep{zellers2019blogpost}. We also find that \modelname~when trained to discriminate between real and fake \modelname-generated news can detect GPT2-Mega generated news as fake with 96\% accuracy.}

\section{How does a model distinguish between human and machine text?}
\begin{figure}[t!]
\centering\small
\begin{minipage}{.54\textwidth}
  \centering\small
  \includegraphics[width=\linewidth]{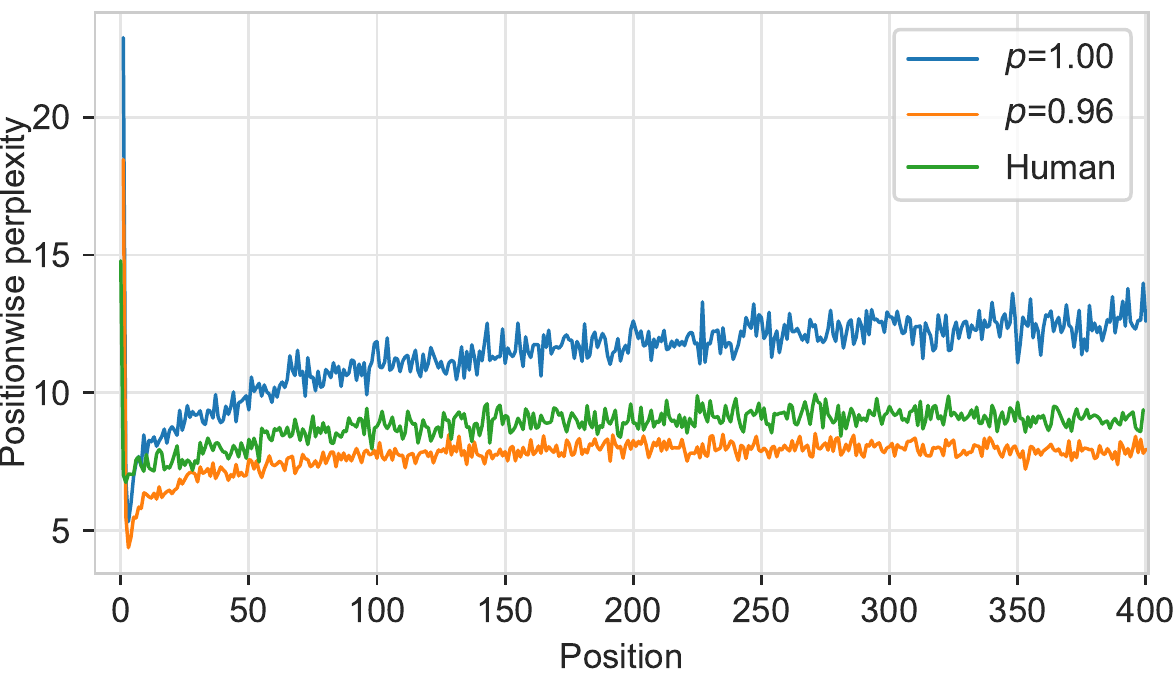}\vspace{-2mm}
  \captionof{figure}{Perplexities of \modelname-Mega, averaged over each position in the \bodyfield~(after conditioning on metadata). We consider human-written with \modelname-Mega generated text at $p{=}1$ (random sampling) and $p{=}.96$. The perplexity of randomly sampled text is higher than human-written text, and the gap increases with position. This suggests that sampling without variance reduction increasingly falls out-of-distribution.}
  \label{fig:pplbypos}
  \vspace{-3mm}
\end{minipage}%
\hspace{.039\textwidth}\begin{minipage}{.42\textwidth}
\vspace{-3mm}
  \centering\small
  \includegraphics[width=\linewidth]{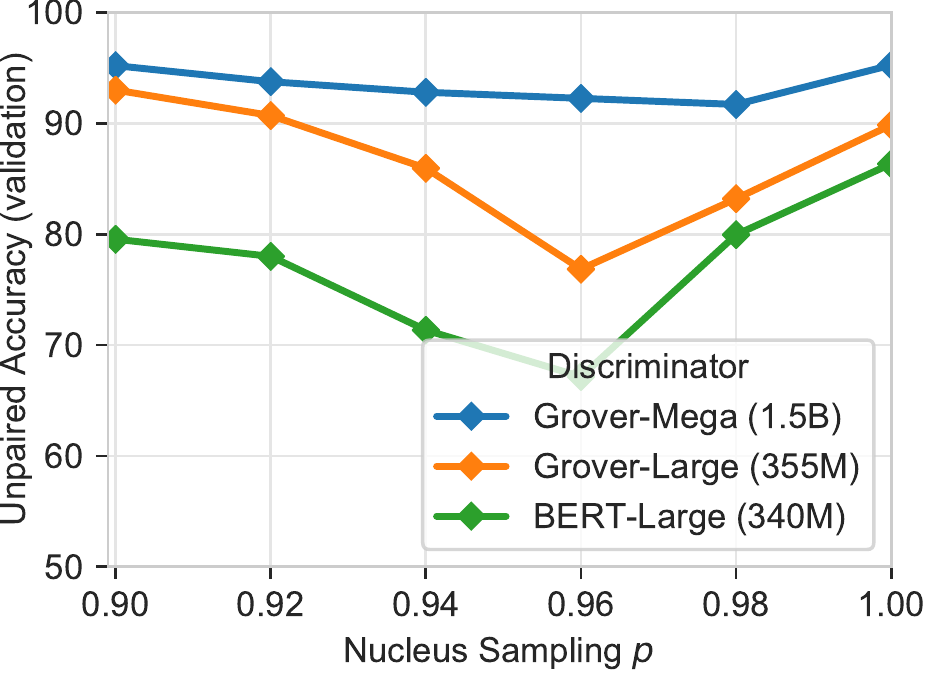}\vspace{-2mm}
  \captionof{figure}{Unpaired validation accuracy, telling apart generated news articles (from \modelname~Mega) from real articles, at different variance reduction thresholds $p$ (for Nucleus Sampling). 
  Results varying $p$ show a sweet spot ($p=0.94$ -- $0.98$) wherein discrimination is hardest.}
  \label{fig:varyingp}
    \vspace{-3mm}
\end{minipage}%
\vspace{-2mm}\end{figure}
\label{sec:analysis}
In this section, we explore why \modelnamefordisc~performs best at detecting fake news generated by other \modelname~models. We find that there is a double-bind between \textbf{exposure bias} and \textbf{variance-reduction} algorithms that alleviate these biases while at the same time creating other artifacts. 

\paragraph{Exposure Bias.} Models maximizing Equation~\ref{eq:lmodeling} are trained only conditioned on human-written text, never on its own generations, creating a problem known as exposure bias \citep{Ranzato2016SequenceLT}. 

We investigate the importance of exposure bias towards creating artifacts. In Figure~\ref{fig:pplbypos} we plot the perplexities given by \modelname-Mega over each position for \bodyfield~text at top-$p$ thresholds of $0.96$ and $1$, as well as over human text. Generating the first token after {\tt\small <start\bodyfield>} results in high perplexity. However, the rest of the positions show a curious pattern: the perplexity of human-written text is lower than randomly sampled text, and this gap increases with sequence length, suggesting that random sampling causes \modelname~to fall increasingly out of the distribution of human language. However, limiting the variance ($p{=}0.96$) lowers the resulting perplexity and limits its growth.

\paragraph{Limiting the variance of a model also creates artifacts}
On the other hand, clipping the model's variance also leaves an artifact, as prior work has observed for top-$k$ sampling \citep{strobelt2019gltr}. A similar phenomenon holds for Nucleus (top-$p$) sampling. The probability of observing a human-written article where all tokens are drawn from the top-$p$\% of the distribution is $p^n$, where $n$ is the document's length. This probability goes to zero as $n$ increases. However, for Nucleus Sampled text -- in which the final $1{-}p$ is cut off -- all tokens come from the top-$p$. 

The visibility of the artifacts depends on the choice of discriminator.
The top-$p$ at each timestep is calculated under the generator's worldview, meaning that if the discriminator models text in a different way, it might have a harder time pinpointing the empty $1{-}p$ tail. This could explain BERT's lower performance during discrimination.

\paragraph{A sweet spot of careful variance reduction}
Not reducing the variance, as well as significantly reducing the variance, both cause problems. Might there be a \emph{sweet spot} for how much to truncate the variance, to make discrimination maximally hard? In Figure~\ref{fig:varyingp}, we show results varying the top-$p$ threshold for the discrimination task applied to \modelname-Mega's generations. The results indeed show a sweet spot, roughly between $p{=}0.94$ and $p{=}0.98$ depending on the discriminator, wherein discrimination is hardest. Interestingly, we note that the most adversarial top-$p$ threshold for BERT-Large is considerably lower than the corresponding top-$p$ for \modelnamefordisc-Large of the same size. This supports our hypothesis that BERT's view of language differs markedly from \modelname; using a lower top-$p$ threshold does not seem to give it much more information about the missing tail. 

\textbf{Overall}, our analysis suggests that \modelname~might be the best at catching \modelname~because it is the best at knowing where the tail is, and thus whether it was truncated.






\section{Conclusion: a Release Strategy for \modelname}
This paper investigates the threats posed by adversaries seeking to spread disinformation. Our sketch of what these threats might look like -- a controllable language model named \modelname~-- suggests that these threats are real and dangerous. \modelname~can rewrite propaganda articles, with humans rating the rewritten versions as more trustworthy. At the same time, there are defenses to these models -- notably, in the form of \modelname~itself. We conclude with a discussion of next steps and ethical considerations.

\paragraph{The Era of Neural Disinformation.} Though training \modelname~was challenging, it is easily achievable by real-world adversaries today. Obtaining the data required through Common Crawl cost \$10k in AWS credits and can be massively parallelized over many CPUs. Training \modelname-Mega is relatively inexpensive: at a cost of \$0.30 per TPU v3 core-hour and two weeks of training, the total cost is \$25k. Spending more money and engineering time could yield even more powerful generators.

\paragraph{Release of generators is critical.} At first, it would seem like keeping models like \modelname~private would make us safer. However, \modelname~serves as an effective detector of neural fake news, even when the generator is much larger (Section~\ref{sec:detection}). If generators are kept private, then there will be little recourse against adversarial attacks. We thus released our models to researchers \citep{zellers2019whywereleasedgrover}.


\paragraph{Future of progress in generation.} 
Models like BERT are strong discriminators for many NLP tasks, but they are not as good at detecting \modelname's generations as left-to-right models like \modelname, even after domain adaptation. One hypothesis is that the artifacts shown in Section~\ref{sec:analysis} are most visible to a left-to-right discriminator. This also suggests that recent progress on generating text in any order \citep{gu2019insertion, stern2019insertion, ghazvininejad2019constant} may lead to models that evade a \modelnamefordisc~discriminator. Likewise, models that are trained conditioned on their own predictions might avoid exposure bias, however, these objectives often lead to low performance on language tasks \citep{caccia2018language}. One additional possibility is the use of Adversarial Filtering \citep{zellers2018swagaf, zellers2018hellaswag} to oversample and then select a subset of generations. However, we found this didn't work well for very long sequences (up to 1024 BPE tokens), possibly as these are far from the `Goldilocks Zone' wherein discrimination is hard for machines.

\paragraph{Additional threat models.} In this paper, we studied the threat model whereby an adversary generates an entire news article from scratch, given minimal context. Other threat models are possible: for instance, an adversary might generate comments or have entire dialogue agents, they might start with a human-written news article and modify a few sentences, and they might fabricate images or video. These threat models ought to be studied by researchers also so that we can create better defenses. 

\paragraph{Machine-generated real news?} Our study focused on detecting machine-written fake news, though the same \modelname~approach can be used for spotting human-written fake news as well \citep{zellers2019blogpost}. However, machines can also generate truthful news using templated systems. Domains with templated news articles exist in our dataset,\footnote{An example is \url{https://americanbankingnews.com}.} and are easy for \modelname~to spoof convincingly. 

\paragraph{Future of progress in discrimination.} Our discriminators are effective, but they primarily leverage distributional features rather than evidence. In contrast, humans assess whether an article is truthful by relying on a model of the world, assessing whether the evidence in the article matches that model.  Future work should investigate integrating knowledge into the discriminator (e.g. for claim verification in FEVER; \citealp{thorne2018fever}).
An open question is to scale progress in this task towards entire news articles, and without paired evidence (similar to open-domain QA; \citealp{chen2017reading}).

\paragraph{What should platforms do?} Video-sharing platforms like YouTube use deep neural networks to scan videos while they are uploaded, to filter out content like pornography \citep{hosseini2017attacking}. We suggest platforms do the same for news articles. An ensemble of deep generative models, such as \modelname, can analyze the content of text -- together with more shallow models that predict human-written disinformation. However, humans must still be in the loop due to dangers of flagging real news as machine-generated, and possible unwanted social biases of these models.

\vspace*{-2mm}
\section*{Acknowledgments}
\vspace*{-2mm}
We thank the anonymous reviewers, as well as Dan Weld, for their helpful feedback. Thanks also to Zak Stone and the Google Cloud TPU team for help with the computing infrastructure. This work was supported by the National Science Foundation through a Graduate Research Fellowship (DGE-1256082) and NSF grants (IIS-1524371, 1637479, 165205, 1703166), the DARPA CwC program through ARO (W911NF-15-1-0543), the Sloan Research Foundation through a Sloan Fellowship, the Allen Institute for Artificial Intelligence, the NVIDIA Artificial Intelligence Lab, Samsung through a Samsung AI research grant, and gifts by Google and Facebook. Computations on beaker.org were supported in part by credits from Google Cloud.

\bibliographystyle{plainnat}
\bibliography{main}

\begin{thebibliography}{47}
\providecommand{\natexlab}[1]{#1}
\providecommand{\url}[1]{\texttt{#1}}
\expandafter\ifx\csname urlstyle\endcsname\relax
  \providecommand{\doi}[1]{doi: #1}\else
  \providecommand{\doi}{doi: \begingroup \urlstyle{rm}\Url}\fi

\bibitem[Bengio et~al.(2003)Bengio, Ducharme, Vincent, and
  Jauvin]{bengio2003neural}
Yoshua Bengio, R{\'e}jean Ducharme, Pascal Vincent, and Christian Jauvin.
\newblock A neural probabilistic language model.
\newblock \emph{Journal of machine learning research}, 3\penalty0
  (Feb):\penalty0 1137--1155, 2003.

\bibitem[Bradshaw and Howard(2017)]{bradshaw2017troops}
Samantha Bradshaw and Philip Howard.
\newblock Troops, trolls and troublemakers: A global inventory of organized
  social media manipulation.
\newblock Technical report, Oxford Internet Institute, 2017.

\bibitem[Caccia et~al.(2018)Caccia, Caccia, Fedus, Larochelle, Pineau, and
  Charlin]{caccia2018language}
Massimo Caccia, Lucas Caccia, William Fedus, Hugo Larochelle, Joelle Pineau,
  and Laurent Charlin.
\newblock Language gans falling short.
\newblock \emph{arXiv preprint arXiv:1811.02549}, 2018.

\bibitem[Chen et~al.(2017)Chen, Fisch, Weston, and Bordes]{chen2017reading}
Danqi Chen, Adam Fisch, Jason Weston, and Antoine Bordes.
\newblock Reading wikipedia to answer open-domain questions.
\newblock In \emph{Proceedings of the 55th Annual Meeting of the Association
  for Computational Linguistics (Volume 1: Long Papers)}, pages 1870--1879,
  2017.

\bibitem[Devlin et~al.(2018)Devlin, Chang, Lee, and Toutanova]{devlin2018bert}
Jacob Devlin, Ming-Wei Chang, Kenton Lee, and Kristina Toutanova.
\newblock Bert: Pre-training of deep bidirectional transformers for language
  understanding.
\newblock \emph{arXiv preprint arXiv:1810.04805}, 2018.

\bibitem[Dicker(2016)]{fakenewslist}
Rachel Dicker.
\newblock {Avoid These Fake News Sites at All Costs}.
\newblock
  \url{https://www.usnews.com/news/national-news/articles/2016-11-14/avoid-these-fake-news-sites-at-all-costs},
  2016.
\newblock [Online; accessed 22-May-2019].

\bibitem[Dwoskin and Romm(2018)]{dwoskin2018facebook}
Elizabeth Dwoskin and Tony Romm.
\newblock Facebook says it has uncovered a coordinated disinformation operation
  ahead of the 2018 midterm elections.
\newblock \emph{The Washington Post}, 2018.

\bibitem[Fan et~al.(2018)Fan, Lewis, and Dauphin]{fan2018hierarchical}
Angela Fan, Mike Lewis, and Yann Dauphin.
\newblock Hierarchical neural story generation.
\newblock In \emph{Proceedings of the 56th Annual Meeting of the Association
  for Computational Linguistics (Volume 1: Long Papers)}, pages 889--898, 2018.

\bibitem[Faris et~al.(2017)Faris, Roberts, Etling, Bourassa, Zuckerman, and
  Benkler]{faris2017partisanship}
Robert Faris, Hal Roberts, Bruce Etling, Nikki Bourassa, Ethan Zuckerman, and
  Yochai Benkler.
\newblock Partisanship, propaganda, and disinformation: Online media and the
  2016 us presidential election.
\newblock \emph{Berkman Klein Center Research Publication 2017-6.}, 2017.

\bibitem[Ghazvininejad et~al.(2019)Ghazvininejad, Levy, Liu, and
  Zettlemoyer]{ghazvininejad2019constant}
Marjan Ghazvininejad, Omer Levy, Yinhan Liu, and Luke Zettlemoyer.
\newblock Constant-time machine translation with conditional masked language
  models.
\newblock \emph{arXiv preprint arXiv:1904.09324}, 2019.

\bibitem[Gu et~al.(2019)Gu, Liu, and Cho]{gu2019insertion}
Jiatao Gu, Qi~Liu, and Kyunghyun Cho.
\newblock Insertion-based decoding with automatically inferred generation
  order.
\newblock \emph{arXiv preprint arXiv:1902.01370}, 2019.

\bibitem[Han and Eisenstein(2019)]{han2019unsupervised}
Xiaochuang Han and Jacob Eisenstein.
\newblock Unsupervised domain adaptation of contextualized embeddings: A case
  study in early modern english.
\newblock \emph{arXiv preprint arXiv:1904.02817}, 2019.

\bibitem[Hashimoto et~al.(2019)Hashimoto, Zhang, and
  Liang]{hashimoto2019unifying}
Tatsunori~B Hashimoto, Hugh Zhang, and Percy Liang.
\newblock Unifying human and statistical evaluation for natural language
  generation.
\newblock \emph{arXiv preprint arXiv:1904.02792}, 2019.

\bibitem[Hecht et~al.(2018)Hecht, Wilcox, Bigham, Sch{\"o}ning, Hoque, Ernnst,
  Bisk, De~Russis, Yarosh, Anjum, Contractor, and Wu]{hecht2018s}
Brent Hecht, Lauren Wilcox, Jeffrey~P. Bigham, Johannes Sch{\"o}ning, Ehsan
  Hoque, Jason Ernnst, Yonatan Bisk, Luigi De~Russis, Lana Yarosh, Bushra
  Anjum, Danish Contractor, and Cathy Wu.
\newblock It's time to do something: Mitigating the negative impacts of
  computing through a change to the peer review process.
\newblock \emph{ACM Future of Computing Blog}, 2018.

\bibitem[Holtzman et~al.(2019)Holtzman, Buys, Forbes, and
  Choi]{holtzman2019curious}
Ari Holtzman, Jan Buys, Maxwell Forbes, and Yejin Choi.
\newblock The curious case of neural text degeneration.
\newblock \emph{arXiv preprint arXiv:1904.09751}, 2019.

\bibitem[Hosseini et~al.(2017)Hosseini, Xiao, Clark, and
  Poovendran]{hosseini2017attacking}
Hossein Hosseini, Baicen Xiao, Andrew Clark, and Radha Poovendran.
\newblock Attacking automatic video analysis algorithms: A case study of google
  cloud video intelligence api.
\newblock In \emph{Proceedings of the 2017 on Multimedia Privacy and Security},
  pages 21--32. ACM, 2017.

\bibitem[Hu et~al.(2017)Hu, Yang, Liang, Salakhutdinov, and Xing]{hu2017toward}
Zhiting Hu, Zichao Yang, Xiaodan Liang, Ruslan Salakhutdinov, and Eric~P Xing.
\newblock Toward controlled generation of text.
\newblock In \emph{Proceedings of the 34th International Conference on Machine
  Learning-Volume 70}, pages 1587--1596. JMLR. org, 2017.

\bibitem[Joulin et~al.(2017)Joulin, Grave, Bojanowski, and
  Mikolov]{joulin2017bag}
Armand Joulin, Edouard Grave, Piotr Bojanowski, and Tomas Mikolov.
\newblock Bag of tricks for efficient text classification.
\newblock In \emph{Proceedings of the 15th Conference of the European Chapter
  of the Association for Computational Linguistics: Volume 2, Short Papers},
  volume~2, pages 427--431, 2017.

\bibitem[J{\'o}zefowicz et~al.(2016)J{\'o}zefowicz, Vinyals, Schuster, Shazeer,
  and Wu]{Jzefowicz2016ExploringTL}
Rafal J{\'o}zefowicz, Oriol Vinyals, Mike Schuster, Noam Shazeer, and Yonghui
  Wu.
\newblock Exploring the limits of language modeling.
\newblock \emph{CoRR}, abs/1602.02410, 2016.

\bibitem[Kingma and Ba(2014)]{Kingma2014AdamAM}
Diederik~P. Kingma and Jimmy Ba.
\newblock Adam: A method for stochastic optimization.
\newblock \emph{CoRR}, abs/1412.6980, 2014.

\bibitem[Melford and Fagan(2019)]{melford2019disinfo}
Clare Melford and Craig Fagan.
\newblock Cutting the funding of disinformation: The ad-tech solution.
\newblock Technical report, The Global Disinformation Index, 2019.

\bibitem[Mosseri(2018)]{mosseri2018news}
Adam Mosseri.
\newblock News feed fyi: Helping ensure news on facebook is from trusted
  sources.
\newblock \emph{Facebook Newsroom}, 19, 2018.

\bibitem[Ott et~al.(2011)Ott, Choi, Cardie, and Hancock]{ott2011finding}
Myle Ott, Yejin Choi, Claire Cardie, and Jeffrey~T Hancock.
\newblock Finding deceptive opinion spam by any stretch of the imagination.
\newblock In \emph{Proceedings of the 49th annual meeting of the association
  for computational linguistics: Human language technologies-volume 1}, pages
  309--319. Association for Computational Linguistics, 2011.

\bibitem[P{\'e}rez-Rosas et~al.(2018)P{\'e}rez-Rosas, Kleinberg, Lefevre, and
  Mihalcea]{perez2018automatic}
Ver{\'o}nica P{\'e}rez-Rosas, Bennett Kleinberg, Alexandra Lefevre, and Rada
  Mihalcea.
\newblock Automatic detection of fake news.
\newblock In \emph{Proceedings of the 27th International Conference on
  Computational Linguistics}, pages 3391--3401, 2018.

\bibitem[Peters et~al.(2018)Peters, Neumann, Iyyer, Gardner, Clark, Lee, and
  Zettlemoyer]{peters2018deep}
Matthew Peters, Mark Neumann, Mohit Iyyer, Matt Gardner, Christopher Clark,
  Kenton Lee, and Luke Zettlemoyer.
\newblock Deep contextualized word representations.
\newblock In \emph{Proceedings of the 2018 Conference of the North American
  Chapter of the Association for Computational Linguistics: Human Language
  Technologies, Volume 1 (Long Papers)}, volume~1, pages 2227--2237, 2018.

\bibitem[Radford et~al.(2018)Radford, Narasimhan, Salimans, and
  Sutskever]{radford2018improving}
Alec Radford, Karthik Narasimhan, Tim Salimans, and Ilya Sutskever.
\newblock Improving language understanding by generative pre-training.
\newblock Technical report, OpenAI, 2018.
\newblock URL \url{https://blog.openai.com/language-unsupervised/}.

\bibitem[Radford et~al.(2019)Radford, Wu, Child, Luan, Amodei, and
  Sutskever]{radford2019gpttwo}
Alec Radford, Jeffrey Wu, Rewon Child, David Luan, Dario Amodei, and Ilya
  Sutskever.
\newblock Language models are unsupervised multitask learners.
\newblock Technical report, OpenAI, 2019.

\bibitem[Ranzato et~al.(2016)Ranzato, Chopra, Auli, and
  Zaremba]{Ranzato2016SequenceLT}
Marc'Aurelio Ranzato, Sumit Chopra, Michael Auli, and Wojciech Zaremba.
\newblock Sequence level training with recurrent neural networks.
\newblock In \emph{ICLR}. ICLR, 2016.

\bibitem[Rashkin et~al.(2017)Rashkin, Choi, Jang, Volkova, and
  Choi]{rashkin2017truth}
Hannah Rashkin, Eunsol Choi, Jin~Yea Jang, Svitlana Volkova, and Yejin Choi.
\newblock Truth of varying shades: Analyzing language in fake news and
  political fact-checking.
\newblock In \emph{Proceedings of the 2017 Conference on Empirical Methods in
  Natural Language Processing}, pages 2931--2937, 2017.

\bibitem[Shao et~al.(2016)Shao, Ciampaglia, Flammini, and
  Menczer]{shao2016hoaxy}
Chengcheng Shao, Giovanni~Luca Ciampaglia, Alessandro Flammini, and Filippo
  Menczer.
\newblock Hoaxy: A platform for tracking online misinformation.
\newblock In \emph{Proceedings of the 25th international conference companion
  on world wide web}, pages 745--750. International World Wide Web Conferences
  Steering Committee, 2016.

\bibitem[Shazeer and Stern(2018)]{shazeer2018adafactor}
Noam Shazeer and Mitchell Stern.
\newblock Adafactor: Adaptive learning rates with sublinear memory cost.
\newblock In \emph{International Conference on Machine Learning}, pages
  4603--4611, 2018.

\bibitem[Solaiman et~al.(2019)Solaiman, Brundage, Clark, Askell, Herbert-Voss,
  Wu, Radford, and Wang]{solaiman2019release}
Irene Solaiman, Miles Brundage, Jack Clark, Amanda Askell, Ariel Herbert-Voss,
  Jeff Wu, Alec Radford, and Jasmine Wang.
\newblock Release strategies and the social impacts of language models.
\newblock \emph{arXiv preprint arXiv:1908.09203}, 2019.

\bibitem[Stern et~al.(2019)Stern, Chan, Kiros, and
  Uszkoreit]{stern2019insertion}
Mitchell Stern, William Chan, Jamie Kiros, and Jakob Uszkoreit.
\newblock Insertion transformer: Flexible sequence generation via insertion
  operations.
\newblock \emph{arXiv preprint arXiv:1902.03249}, 2019.

\bibitem[Strobelt and Gehrmann(2019)]{strobelt2019gltr}
Hendrik Strobelt and Sebastian Gehrmann.
\newblock Catching a unicorn with gltr: A tool to detect automatically
  generated text.
\newblock Technical report, Harvard, 2019.

\bibitem[Swire et~al.(2017)Swire, Ecker, and Lewandowsky]{swire2017role}
Briony Swire, Ullrich~KH Ecker, and Stephan Lewandowsky.
\newblock The role of familiarity in correcting inaccurate information.
\newblock \emph{Journal of experimental psychology: learning, memory, and
  cognition}, 43\penalty0 (12):\penalty0 1948, 2017.

\bibitem[Thorne et~al.(2018)Thorne, Vlachos, Christodoulopoulos, and
  Mittal]{thorne2018fever}
James Thorne, Andreas Vlachos, Christos Christodoulopoulos, and Arpit Mittal.
\newblock Fever: a large-scale dataset for fact extraction and verification.
\newblock In \emph{Proceedings of the 2018 Conference of the North American
  Chapter of the Association for Computational Linguistics: Human Language
  Technologies, Volume 1 (Long Papers)}, pages 809--819, 2018.

\bibitem[Vargo et~al.(2018)Vargo, Guo, and Amazeen]{vargo2018agenda}
Chris~J Vargo, Lei Guo, and Michelle~A Amazeen.
\newblock The agenda-setting power of fake news: A big data analysis of the
  online media landscape from 2014 to 2016.
\newblock \emph{New Media \& Society}, 20\penalty0 (5):\penalty0 2028--2049,
  2018.

\bibitem[Vaswani et~al.(2017)Vaswani, Shazeer, Parmar, Uszkoreit, Jones, Gomez,
  Kaiser, and Polosukhin]{vaswani2017attention}
Ashish Vaswani, Noam Shazeer, Niki Parmar, Jakob Uszkoreit, Llion Jones,
  Aidan~N Gomez, {\L}ukasz Kaiser, and Illia Polosukhin.
\newblock Attention is all you need.
\newblock In \emph{Proceedings of the 31st International Conference on Neural
  Information Processing Systems}, pages 6000--6010. Curran Associates Inc.,
  2017.

\bibitem[Wang et~al.(2018)Wang, Singh, Michael, Hill, Levy, and
  Bowman]{wang2018glue}
Alex Wang, Amapreet Singh, Julian Michael, Felix Hill, Omer Levy, and Samuel~R
  Bowman.
\newblock Glue: A multi-task benchmark and analysis platform for natural
  language understanding.
\newblock \emph{arXiv preprint arXiv:1804.07461}, 2018.

\bibitem[Wang(2017)]{wang2017liar}
William~Yang Wang.
\newblock “liar, liar pants on fire”: A new benchmark dataset for fake news
  detection.
\newblock In \emph{Proceedings of the 55th Annual Meeting of the Association
  for Computational Linguistics (Volume 2: Short Papers)}, pages 422--426,
  2017.

\bibitem[Wardle(2017)]{wardle2017fake}
Claire Wardle.
\newblock Fake news. it’s complicated.
\newblock \emph{First Draft News}, 16, 2017.

\bibitem[Wardle and Derakhshan(2017)]{wardle2017information}
Claire Wardle and Hossein Derakhshan.
\newblock Information disorder: Toward an interdisciplinary framework for
  research and policy making.
\newblock \emph{Council of Europe report, DGI (2017)}, 9, 2017.

\bibitem[Zellers(2019)]{zellers2019whywereleasedgrover}
Rowan Zellers.
\newblock Why we released grover.
\newblock Technical report, 2019.
\newblock URL \url{https://thegradient.pub/why-we-released-grover/}.

\bibitem[Zellers et~al.(2018)Zellers, Bisk, Schwartz, and
  Choi]{zellers2018swagaf}
Rowan Zellers, Yonatan Bisk, Roy Schwartz, and Yejin Choi.
\newblock Swag: A large-scale adversarial dataset for grounded commonsense
  inference.
\newblock In \emph{Proceedings of the 2018 Conference on Empirical Methods in
  Natural Language Processing (EMNLP)}, 2018.

\bibitem[Zellers et~al.(2019{\natexlab{a}})Zellers, Bisk, Farhadi, and
  Choi]{zellers2019vcr}
Rowan Zellers, Yonatan Bisk, Ali Farhadi, and Yejin Choi.
\newblock From recognition to cognition: Visual commonsense reasoning.
\newblock In \emph{The IEEE Conference on Computer Vision and Pattern
  Recognition (CVPR)}, 2019{\natexlab{a}}.

\bibitem[Zellers et~al.(2019{\natexlab{b}})Zellers, Holtzman, Bisk, Farhadi,
  and Choi]{zellers2018hellaswag}
Rowan Zellers, Ari Holtzman, Yonatan Bisk, Ali Farhadi, and Yejin Choi.
\newblock Hellaswag: Can a machine really finish your sentence?
\newblock In \emph{Proceedings of the 57th Annual Meeting of the Association
  for Computational Linguistics}, 2019{\natexlab{b}}.

\bibitem[Zellers et~al.(2019{\natexlab{c}})Zellers, Holtzman, Rashkin, Bisk,
  Farhadi, Roesner, and Choi]{zellers2019blogpost}
Rowan Zellers, Ari Holtzman, Hannah Rashkin, Yonatan Bisk, Ali Farhadi,
  Franziska Roesner, and Yejin Choi.
\newblock Counteracting neural disinformation with grover.
\newblock Technical report, 2019{\natexlab{c}}.
\newblock URL
  \url{https://medium.com/ai2-blog/counteracting-neural-disinformation-with-grover-6cf6690d463b}.

\end{thebibliography}

\appendix
\clearpage
\section*{Supplemental Material}
\section{Optimization Hyperparameters}
\label{sec:optimizationhyperparameters}
For our input representation, we use the same BPE vocabulary as \citep{radford2019gpttwo}. We use Adafactor \citep{shazeer2018adafactor} as our optimizer. Common optimizers such as Adam \citep{Kingma2014AdamAM} tend to work well, but the memory cost scales linearly with the number of parameters, which renders training \modelname-Mega all but impossible. Adafactor alleviates this problem by factoring the second-order momentum parameters into a tensor product of two vectors. We used a maximum learning rate of 1e-4 with linear warm-up over the first 10,000 iterations, and decay over the remaining iterations. We set Adafactor's $\beta_1=0.999$ and clipped updates for each parameter to a root-mean-squared of at most 1. Last, we applied weight decay with coefficient $0.01$. We used a batch size of 512 on 256 TPU v3 cores. which corresponds to roughly 20 epochs through our news dataset. The total training time required roughly two weeks.

\section{Real News and Propaganda Websites}
\label{sec:newssites}
\newcommand\fnl[1]{{\tt\small \href{https://#1}{#1}}}
In our generation experiments (Section~\ref{sec:genexps}), we consider a set of mainstream as well as propaganda websites. We used the following websites as `real news': \fnl{theguardian.com}, \fnl{reuters.com}, 
\fnl{nytimes.com},
\fnl{theatlantic.com},
\fnl{usatoday.com},
\fnl{huffingtonpost.com},
and 
\fnl{nbcnews.com}. 
For propaganda sites, we chose sites that have notably spread misinformation \citep{fakenewslist} or propaganda\footnote{For more information, see the Media Bias Chart at {\tt\scriptsize \href{https://www.adfontesmedia.com/}{adfontesmedia.com/}}.}. These were \fnl{breitbart.com}, \fnl{infowars.com}, \fnl{wnd.com}, \fnl{bigleaguepolitics.com}, and \fnl{naturalnews.com}.

\section{Domain Adaptation of BERT}
\label{sec:bertda}
BERT \citep{devlin2018bert} is a strong model for most classification tasks. However, care must be taken to format the input in the right way, particularly because BERT is pretrained in a setting where it is given two spans (separated by a special {\small\tt [SEP]} token). We thus use the following input format. The first span consists of the metadata, with each field prefixed by its name in brackets (e.g. `{\small\tt [title]}'). The second span consists of the body. Because the generations are cased (with capital and lowercase letters), we used the `cased' version of BERT.

Past work (e.g. \cite{zellers2019vcr, han2019unsupervised}) has found that BERT, like other language models, benefits greatly from domain adaptation.  We thus perform domain adaptation on BERT, adapting it to the news domain, by training it on \datasetname~for 50k iterations at a batch size of 256. Additionally, BERT was trained with a sequence length of at most 512 WordPiece tokens, but generations from \modelname~are much longer (1024 BPE tokens). Thus, we initialized new position embeddings for positions 513-1024, and performed domain adaptation at a length of 1024 WordPiece tokens.

\section{Hyperparameters for the Discriminators}
\label{sec:dischyperparams}
For our discrimination experiments, we limited the lengths of generations (and human-written articles) to 1024 BPE tokens. This was needed because our discriminators only handle documents up to 1024 words. However, we also found that the longer length empirically discrimination easier for models (see Section~\ref{sec:analysis}).

For our discrimination experiments, we used different hyperparameters depending on the model, after an initial grid search. For BERT, we used the Adam \citep{Kingma2014AdamAM} optimizer with a learning rate of $2e-5$ and a batch size of 64. We trained BERT models for 5 epochs, with a linear warm-up of the learning rate over the initial 20\% iterations. For GPT2 and \modelname, we used the Adam actor optimizer \citep{shazeer2018adafactor} optimizer with a learning rate of $2e-5$ for all models, and a batch size of 64. We applied an auxiliary language modeling loss for these models with a coefficient of $0.5$. These models were trained for 10 epochs, with a linear warm-up over the initial 20\% iterations. 

\section{Human Evaluation Prompt}
\label{sec:humanevalprompts}
\subsection{Evaluating Quality}
For evaluating the quality of \modelname-written versus human-written news articles, we asked workers the following questions (shown exactly). The answer choices are shown next to the rating under our 1-3 Likert scale (3 being the best, 1 being the worst for each attribute).
\begin{enumerate}[label=(\alph*)]
    \item (Style) Is the style of this article consistent?
    \begin{enumerate}
        \item[3.] \textbf{Yes}, this sounds like an article I would find at an online news source.
        \item[2.] \textbf{Sort of}, but there are certain sentences that are awkward or strange.
        \item[1.] \textbf{No}, it reads like it's written by a madman.
    \end{enumerate}
    \item (Content) Does the content of this article make sense?
        \begin{enumerate}
        \item[3.]  \textbf{Yes}, this article reads coherently.
        \item[2.]  \textbf{Sort of}, but I don't understand what the author means in certain places.
        \item[1.]  \textbf{No}, I have no (or almost no) idea what the author is trying to say.
    \end{enumerate}
    \item (Overall) Does the article read like it comes from a trustworthy source?
    \begin{enumerate}
        \item[3.]  \textbf{Yes}, I feel that this article could come from a news source I would trust.
        \item[2.]  \textbf{Sort of}, but something seems a bit fishy.
        \item[1.]  \textbf{No}, this seems like it comes from an unreliable source.
    \end{enumerate}
\end{enumerate}

\subsection{Evaluating consistency}
To measure consistency between the article and the metadata, we asked the following questions:
\begin{enumerate}[label=(\alph*)]
    \item (Headline) How well does the article body match the following headline? [headline]
        \begin{enumerate}
        \item[3.]  \textbf{Yes}, the article makes sense as something that I would see given the headline.
        \item[2.]  \textbf{Sort of}, the article is somewhat related to the headline, but seems slightly off.
        \item[1.]  \textbf{No}, the article is completely off-topic.
    \end{enumerate}
    \item (Authors) How well does the article body match the following author(s)? [authors]
        \begin{enumerate}
        \item[3.]  \textbf{Yes}, the article makes sense as something that could be written by the author(s).
        \item[2.]  \textbf{Sort of}, the article might have been written by the author(s) above, but it sounds unlikely.
        \item[1.]  \textbf{No}, the article body contains information that says it was written by someone else.
    \end{enumerate}
    \item (Date) How well does the article body match the following date? [date]
        \begin{enumerate}
        \item[3.]  \textbf{Yes}, the article makes sense as something that could have been written on [date].
        \item[2.]  \textbf{Sort of}, the article might have been written on [date], but it sounds unlikely.
        \item[1.]  \textbf{No}, there's information in the article that conflicts the proposed date.
    \end{enumerate}
\end{enumerate}



\section{Examples}
\label{sec:suppexamples}
\begin{figure}
    \centering
\includegraphics[trim={0 3cm 0 0},clip,width=\columnwidth]{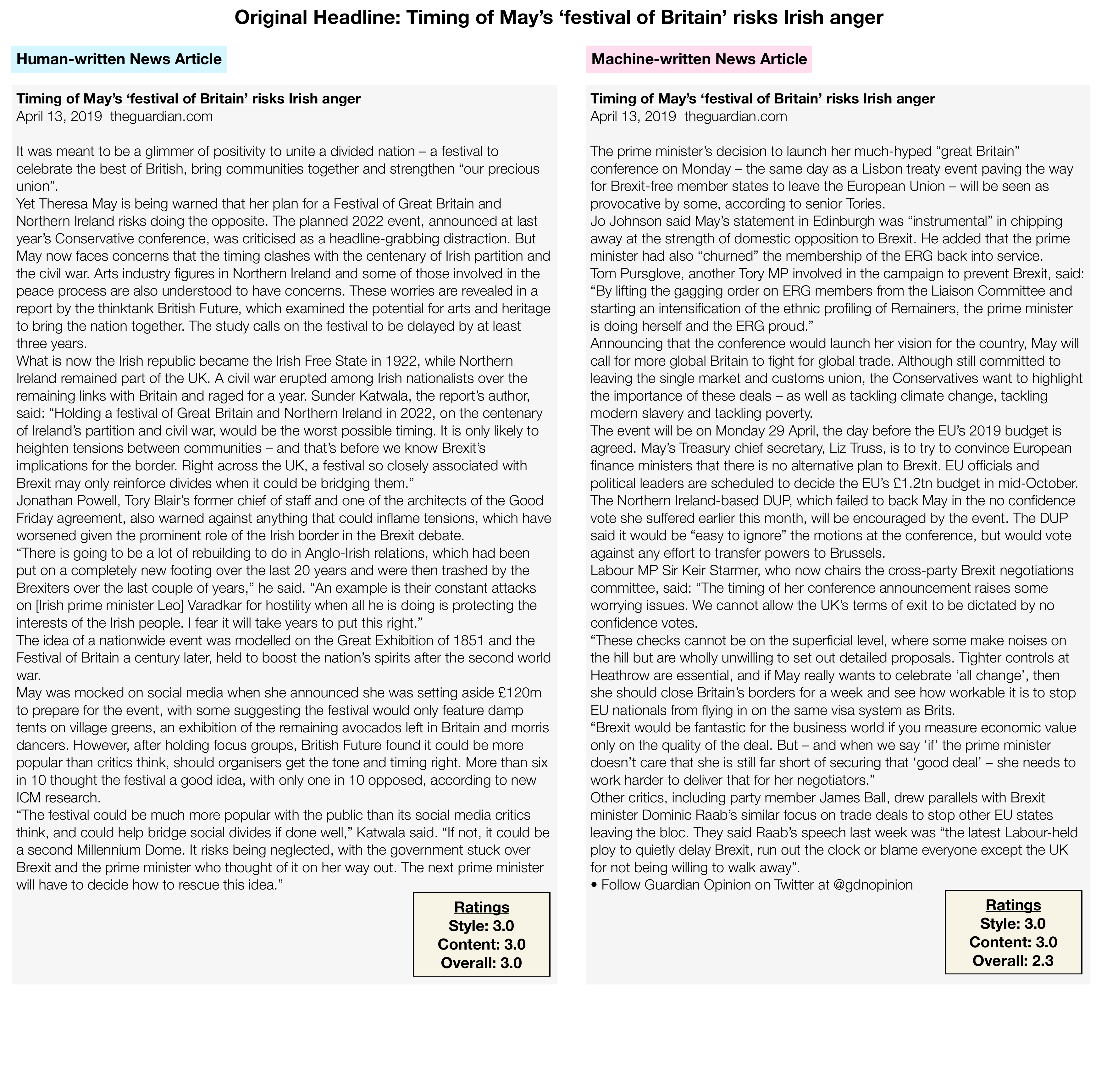}
    \caption{Example of human-written news and machine-written news articles about the same headline from The Guardian with the average ratings from human rating study.}
    \label{fig:realex}
\end{figure}

\begin{figure}
    \centering
\includegraphics[width=\columnwidth]{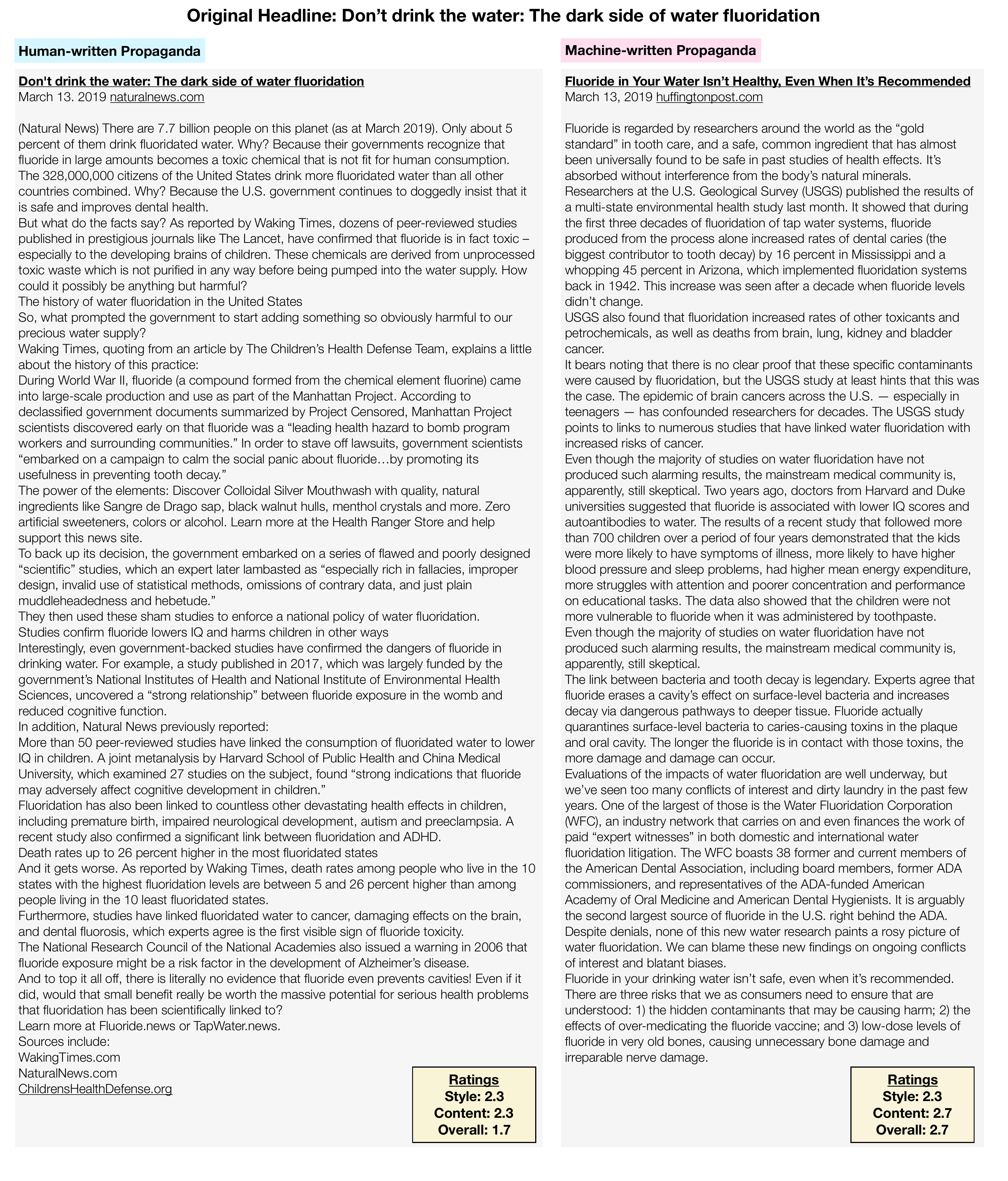}
    \caption{Example of human-written and machine-written articles arguing against fluoride with the average ratings from human rating study.}
    \label{fig:propex}
\end{figure}

In Figures~\ref{fig:realex} and \ref{fig:propex}, we include examples of articles with the average scores given by human raters, who were asked to evaluate the style, content, and overall trustworthiness.  In Figure~\ref{fig:realex}, we show a real article ({\tt\small Human News}) posted by the Guardian along with an article from \modelname~ ({\tt\small Machine News}) made using the same metadata.   Figure~\ref{fig:propex} shows a real propaganda article from the Natural News ({\tt\small Human Propaganda}) and an article made with \modelname~ ({\tt\small Machine Propaganda}) with the original headline and the style of Huffington Post (\modelname~ was used to re-write the title to be more stylistically similar to the Huffington Post, as well).

We also present several other generated examples, generated from \modelname-Mega with a top-$p$ threshold of $p{=}0.95$. All of the examples are cut off to 1024 generated BPE tokens, since this is our setup for discrimination. 
\begin{enumerate}[wide, leftmargin=10pt, labelwidth=!,labelindent=-2pt,itemsep=1pt,topsep=0pt,label=\textbf{\alph*}.]
    \item \modelname~can generate controlled propaganda. In Figure~\ref{fig:teaserfigurecontinued}, we show the continuation from Figure~\ref{fig:teaser}, about a link found between autism and vaccines.
    \item \modelname~can spoof the identity of writers. In Figure~\ref{fig:paulkrugman} we show a realistic-looking editorial seemingly from New York Times columnist Paul Krugman.
    \item \modelname~can generate fake political news. In Figure~\ref{fig:trumpimpeached} we show an article generated about Trump being impeached, written in the style of the Washington Post.
    \item \modelname~can generate fake movie reviews (opinion spam; \cite{ott2011finding}). In Figure~\ref{fig:sharknado} we show a movie review, generated in the style of LA Times Movie Critic Kenneth Turan, for Sharknado 6, `The Last Sharknado: It's About Time'
    \item \modelname~can generate fake business news. In Figure~\ref{fig:uberfordogs}, we show an article generated about an `Uber for Dogs' startup.
\end{enumerate}

\begin{figure}[t]
Provided \headlinefield: `Link found between autism and vaccines'
\hrule
Generated \bodyfield:
{\small
Those who have been vaccinated against measles have a more than 5-fold higher chance of developing autism, researchers at the University of California San Diego School of Medicine and the Centers for Disease Control and Prevention report today in the Journal of Epidemiology and Community Health.

Those who have been vaccinated against measles have a more than 5-fold higher chance of developing autism.

Earlier studies had linked past vaccines with the condition, but it was not clear whether those links were causal or indirect. Among research groups, the UCSD and CDC teams used new methods that enabled them to determine the associations between the virus and the disease.

While researchers in the previous studies had observed associations between those vaccinated against measles and autism in children whose mothers had received the virus, the current UCSD/CDC team focused on childhood vaccines given during the first two years of life because it is commonly believed those vaccines begin to clear the body’s immune system before the onset of autism. The UCSD/CDC team compared children who had been vaccinated against measles, mumps and rubella (MMR) and those who had not.

“Because this is the first study to focus on MMR-vaccine syndrome, it was also the first time that this syndrome had been scientifically characterized,” said study lead author William Chan, MD, MPH, a researcher in the Department of Pediatrics in the UC San Diego School of Medicine. “We were able to identify certain subgroups of those vaccinated against MMR and compare them to those not vaccinated.”

Mumps, measles and rubella vaccines are administered in either three or five doses, depending on the age. A recent study by the Public Health Agency of Canada found only 10 percent of children vaccinated against measles had a developing autism disorder when compared to 73 percent of children vaccinated with one dose, 50 percent of those vaccinated with two doses and 45 percent of those vaccinated with three doses. Those results do not surprise the UC San Diego researchers, who found those vaccinated with three or more doses were at an additional 5- to 8-fold increased risk.

The current UCSD/CDC analysis followed the same years of data as earlier studies, but used new methods to accurately pinpoint the timing of those immunity-shifting epidemics. For example, previous studies showed that autism was on the rise in 1995-96, but never topped 50 cases until 2001-02, when measles overtook mumps as the leading cause of infant death due to disease. In 2003, several measles outbreaks were triggered by parents who had ignored warnings about the vaccine.

“Our research was critical in providing details of when outbreaks occurred in regions where there was significant vaccination efforts,” said study co-author Louis Cooley, MD, PhD, professor of Pediatrics and professor of Communicable Diseases and Epidemiology at UCSD School of Medicine. “We were able to show that if you get your MMR vaccine in early childhood, you should reduce your risk of autism by 33 percent, with every passing year bringing a slightly lower risk.”

Researchers found that low vaccination rates — 42 percent in 2003-04 and 75 percent in 2007-08 — and more frequent, and severe, cases of measles predicted higher rates of autism in later years. In 2004-05, the annual rate of autism and other disorders — including intellectual disability — increased by 20 percent, far more than the 5- to 7-fold increase in their model predicted, suggesting there may be a more subtle link between vaccine exposure and autism than previously understood.
Preliminary findings from a separate analysis of the 2006-07 data suggest that autism risk spikes occurred during times when children had fewer routine vaccinations and received older doses of vaccines.

Autism spectrum disorders include autism spectrum disorders (ASD), a group of neurodevelopmental disorders that can occur with or without symptoms of the immune system disorder, including Asperger’s syndrome, pervasive developmental disorder not otherwise specified (PDD-NOS) and pervasive developmental disorder-not otherwise specified (PDD-ENG).

The study was supported by the National Institutes of Health’s National Institute of Child Health and Human Development (CDER), the NIH Program on Smoking and Health and the San Diego Veterans Affairs (SDVA) Healthcare System.

About the University of California, San Diego School of Medicine

Founded in 1897, the School of Medicine has been recognized nationally for outstanding achievement in educating clinical and research physicians. Nobel laureates and more than 170 physicians who practice in the United States and abroad are the graduates of UC San Diego. The school has been recognized as the first medical school in the world to offer its entire physician-practitioner curriculum online, currently serving more than 1,000 medical students worldwide. The school also trains the next generation of scientists, physicians and public health leaders and ranks first in the nation for its graduates’ ability to train other healthcare professionals. UC San Diego is associated with UC San Diego Health System, which has a faculty of more than 9,000 and an annual
} \\

\hrule
\caption{The rest of the example from the teaser figure, about a hypothetical link found between autism and vaccines.}
\label{fig:teaserfigurecontinued}
\end{figure}

\begin{figure}[t]
Provided \domainfield: {\tt\small \href{https://nytimes.com}{nytimes.com}} \\
Provided \authorfield: Paul Krugman \\
Provided \datefield: May 23, 2019

\hrule
Generated \headlinefield: Why Trump could get impeached \\
Generated \bodyfield:
{\small
President Donald Trump could get impeached. Or he could get fired. Or he could be indicted for special-counsel matters involving Stormy Daniels, or the Russians, or his own campaign. But, at the end of the day, what would matter is what House Speaker Nancy Pelosi decides. And here’s why:

First, Pelosi seems open to the idea of impeachment, to the point where last week she endorsed the use of the word in a speech. And this is not to claim that “impeachment” is a possible solution to the current impasse on immigration. Nobody seriously thinks that. Rather, it’s to say that the dividing line between Trump’s impeachment and, on the other hand, the notion that he’s being treated adequately in his courtship of the Republican base, is soft. If Trump runs a replay of his 2016 campaign, the focus on immigration will fade as an issue. Then his campaign’s fiery rhetoric on the economy and taxes will dominate, and get bigger. But as we pointed out yesterday, the “soul” of the Trump campaign was white nationalist sloganeering and authoritarianism. It seems unlikely that this will change.

If it doesn’t change, impeachment seems like the only way to constrain him.

Yet the power of the impeachment path is out of Trump’s hands. The recent passage of the Defense Authorization Act, to cite just one example, which codifies within the law the requirement that presidents get an honest break from investigations, seems to have guaranteed that Bob Mueller will be working hard into the fall. And “executive privilege” — the right of a president to withhold certain materials from the government — is clearly limited by the extent to which federal judges have looked at and rejected it.

So impeachment is less a decision than a historical fact: If you think the election of a president who promised to enrich himself through corruption has happened, a court-ordered investigation, or both, is probably necessary to correct the damage. But it can’t happen while he’s president, or until he’s gone. Even then, there are a few ways that a president can claim executive privilege, or whatever other arguments he can throw out to minimize the pressure on him.

So Congress must seize the moment. The first and most urgent must be an investigation of the Trump campaign’s possible ties to the Russians, and the subsequent question of whether the president knew what was going on. (For full coverage of the Russia story, see here.) The second, more abstract but no less important issue is whether the president’s own actions constitute obstruction of justice. And third, Trump needs to be impeached because his rhetoric has helped put members of his own party who could back his impeachment into near paralysis.

If Republicans don’t take a bold stand, and appoint a special prosecutor to investigate all this — just as Democrats did during Watergate — then their caucus will become dependent on impeachment proceedings to hold its seats. While that might be a good outcome in the short run, it could also signal a likely government shutdown — just as it signaled the risk of a government shutdown during the Clinton impeachment. So it’s a bad decision for the country, or at least for the country’s self-interest.

That said, it could be an even worse one for Trump. True, other presidents have been impeached in part for behavior far more extreme than Trump’s — for cases like Bill Clinton’s loveless marriage, and, more frequently, for their handling of women in private. But never before has the impeachment authority come at the point where a man at the pinnacle of power and influence has cozied up to a hostile foreign power, and openly attacked not just the judiciary, but the rule of law itself.

For those who would recommend removal now for that alone, it’s worth noting that on a serious level, that kind of behavior isn’t akin to the brinkmanship that some U.S. presidents have brought to bear during a confrontation with Cuba. It’s as close as Trump has come to laying out some of the first lines of battle for our uncertain and uneasy world.

And to me, that is a form of obstruction of justice.
} \\
\hrule
\caption{An example of spoofing someone's identity using \modelname-Mega. We generate a convincing New York Times Op-Ed that is seemingly from columnist Paul Krugman.}
\label{fig:paulkrugman}
\end{figure}

\begin{figure}[t]
Provided \domainfield: {\tt\small \href{https://washingtonpost.com}{washingtonpost.com}} \\
Provided \datefield: May 29, 2019 \\
Provided \headlinefield: US House Votes to Impeach Donald Trump After New Evidence Released from Mueller Report

\hrule
Generated \bodyfield:
{\small
WASHINGTON — The House voted to impeach President Donald Trump Wednesday after releasing hundreds of pages of text messages that point to clear evidence of obstruction of justice and communication with the head of the Trump Organization about a potential business deal in Russia.

The 220-197 vote came after weeks of debate over whether new evidence released by special counsel Robert Mueller’s office signaled sufficient grounds for Trump’s removal from office. The president personally denounced the move, announcing his intent to veto the resolution and accusing Democrats of plotting to remove him from office through a “con job.”

“This is all a big con job,” he said during an appearance in Texas. “They got caught.”

A number of Democratic lawmakers went further, calling for Trump’s impeachment on the grounds that he obstructed justice. Democrats hold the House majority, and if the Senate fails to convict Trump, the president can be removed from office by a two-thirds vote.

But even with Trump’s political allies organizing from the White House, Republicans in the House stood united behind their leader, noting the historical difficulty of making the case for impeachment.

“This is not a decision for the president,” said House Speaker Paul Ryan, R-Wis. “It’s not one for this body, and I think it’s a dereliction of duty for the minority party to be making these calls.”

Democrats on the House Judiciary Committee on Tuesday said they were not bound by Ryan’s defense. They released an 11-page joint report with Republicans on the House Oversight and Government Reform Committee that took a different tack, emphasizing that Mueller had not decided whether the case warranted impeachment.

Rep. Jerrold Nadler, D-N.Y., the committee’s chairman, told reporters before the vote that he hoped the two reports would “send a clear message to the president that we in this body are not going to stand for obstruction of justice.”

But when asked whether the report would build the case for impeachment, he said: “What’s far more likely to build a case for impeachment are the indictments and evidence produced through the special counsel’s work.”

On Tuesday night, the White House issued a statement proclaiming that the book by a former FBI deputy director, which has been cited by Democrats in support of impeachment, had been “made up and defamatory” to the president.

That statement drew swift and heavy rebukes from a number of Democrats, who said they had found the statement “beyond astonishing.”

The Democratic report, citing redacted text messages sent by former FBI counterintelligence agent Peter Strzok, revealed what the party called “a pattern of bias” and “an affinity for the Russian regime.”

The report did not include a full, uncensored text message Strzok sent to a friend after Trump was elected that said: “I’ll stop it.” That led Democrats to bemoan the language used, questioning what is or is not in the special counsel’s investigative file.

The report also quoted a review of emails and witness testimony to Mueller that said Trump “has obstructed justice in the Russia investigation” by firing FBI Director James Comey, by urging Attorney General Jeff Sessions to recuse himself from the Russia investigation and by urging his son, Donald Trump Jr., to meet with a Kremlin-connected Russian lawyer at Trump Tower in 2016.

“This has never happened in the history of our country,” Nadler said of the president. “No president has ever tried to interfere with an investigation into his own conduct.”

The vote was also led by Democrats who have been outspoken against Trump, including Rep. Al Green, D-Texas, who took out newspaper ads calling for the president’s impeachment. Rep. Maxine Waters, D-Calif., another outspoken Trump critic, said Trump’s ties to Russia were “becoming more and more ominous,” and said “as we go on, we are going to see if we can’t get the articles of impeachment out.”

Rep. Tony Cárdenas, D-Calif., a member of the Judiciary Committee, said many Democrats would be eager to see whether the newly discovered evidence could be used to support impeachment.

“I think it will be the nail in the coffin of the ‘non-con,'” he said.
}

Generated \headlinefield: Congress voted to impeach President Trump Wednesday; the President vowed to veto the resolution. \\

\hrule
\caption{An example of using \modelname-Mega to generate fake political news, written in the style of the Washington Post. Here, it generates a mostly-realistic-looking article about Trump being impeached.}
\label{fig:trumpimpeached}
\end{figure}

\begin{figure}[t]
Provided \domainfield: {\tt\small \href{https://latimes.com}{latimes.com}} \\
Provided \datefield: August 1, 2018 \\
Provided \authorsfield: Kenneth Turan \\
Provided \headlinefield: Review: "The Last Sharknado: It's About Time" is a gem of an action movie

\hrule
Generated \bodyfield:
{\small
Some heroes take a vacation every now and then, even temporarily, a couple of beachfront condos in a movie world they hope, a la Rick James in “The Beautiful Daze,” won’t have to live through.

Unfortunately, that has never happened to one of the most exceptional directors in the business, Anthony \& Joe Russo, so the moment is altogether too good to spoil, at least for us Southern Californians.

In “The Last Sharknado: It’s About Time,” which had its Los Angeles premiere at the Tower Theatre on Tuesday, we can expect the San Diego-born Anthony Russo to somehow make it through seven years without being inside a sharknado.

Oh, he has a couple of other movies during that time, mostly that stupid’d-up New York “Avengers” picture, but they have nothing to do with sharks or serious action-film plotting.

Don’t worry, he lives to fight another day.

Written by Jason Friedberg and Aaron Seltzer (“Vampires Suck”) and directed by John Hlavin (“Code Name Verity”), “The Last Sharknado” is exactly what you would expect: very silly and thrilling.

Once again the Ryan Newman- and Ian Ziering-costumed stars of the YouTube smash “Sharknado” have a lot of fun poking fun at themselves and the silly film business (“I don’t know if being caged for a few days after we get out is too much to ask”) along with helping a once-beleaguered government figure out what to do.

We start on a lake in North Carolina as Newman, now an exotic dancer, does a good-natured impersonation of Seltzer’s Agent Slutz. He’s not a bad actor, though, and by the end of the first reel “it’s clear he’s never a lunatic.”

So Slutz has just been rehired by the Trump Administration, even though the agency has been effectively taken over by the tag team of the Governator and Alex Jones (the viewer is never told why they are being hired).

As Slutz, Newman is working with his deputy, Ziering, the impetuous marine officer no longer needled by his loyalty to his best friend, Tara Reid, who showed up in the third “Sharknado” and is absent this time.

The three are doing their best to fix things, with Newman looking in particular for more information on how to deal with the five remaining sharks in New York who seem to think they should go with the Trump administration.
Svelt’s plotting goes way beyond the call of duty, of course, for despite the presence of Anthony Mackie and Juliette Lewis, for example, there are many disappointments here, especially in the casting of a grouchy Jeff Goldblum as the villain of the piece.

Still, like the original “Sharknado,” all the elements are exactly right, including some fancy effects (stunt coordinator Zak Penn, “The Avengers”) and some deeply, bizarrely funny dialogue (try spending a few days in a sharknado and after a while questioning your sanity).

Knowing that everything is going to be fine, Ziering tells Newman: “Whether we go sharknado-free is on you.” Come again?

Might we welcome it? If by a miracle we’re not sated by our popcorn, that would be the best way to look at it.
kenneth.turan@latimes.com

Follow me on Twitter @KennethTuran

}

Generated \headlinefield: 'The Last Sharknado' returns with the dramatic thriller you expected — and some unexpected laughs

\hrule
\caption{An example of using \modelname-Mega to generate a fake movie review. Here, we spoof LA Times Film Critic Kenneth Turan and generate a positive movie review for `The Last Sharknado: It's About Time,' the sixth installment in the Sharknado series of movies.}
\label{fig:sharknado}
\end{figure}

\begin{figure}[t]
Provided \domainfield: {\tt\small \href{https://techcrunch.com}{techcrunch.com}} \\
Provided \datefield: May 29, 2019 \\
Provided \headlinefield: Rowan Zellers' new startup raises 17M in Series A round for "Uber for Dogs"

\hrule
Generated \bodyfield:
{\small
In today’s competitive eco-system of startups trying to convince consumers to eschew popular apps in favor of their own, you would think that the dog-walking market would be the easiest to crack. Well, it’s not. But at least one startup believes it has a new and clever solution: apply its core offering of mobile-based recommendation engine algorithms to locate dog owners near them with the end goal of making dog walking more efficient.

The company, owned by Australian startup Yfrog, is called DogSpotter. It is co-founded by entrepreneur Rowan Zellers, along with King’s College London-trained dogsman Craig Osbaldoon.

Launched in Melbourne in November, DogSpotter has now launched in Sydney, Canberra, Adelaide, Perth, and is set to be introduced to Brisbane and Hobart in June, with a platform the company expects to be available in all capital cities in Australia this year. In addition, it is now prepared to accept payment via credit cards in Australia and New Zealand.

In exchange for its name recognition and the unique proposition that it is aiming to turn into a software-as-a-service, DogSpotter has raised \$17 million in a Series A round of funding led by Plug and Play Tech Center, with participation from Open Ocean Capital and previous investors, including the Cahill Foundation.

So what does DogSpotter offer, beyond a fairly basic service that could theoretically be used by anyone, which is at the base of a human-side algorithm that determines the best potential location for a dog owner’s pet — a problem that dogs are notoriously impatient for in cities where dog walking is often unprofitable or a lost art for many?

Well, DogSpotter leverages its core recommendation engine to process the thousands of now readily available reviews of dog care providers via its app that users can find by searching for their own city or a curated selection of recommendations. That way, it’s much quicker than traditional alternatives, where you have to wade through a fair number of reviews to find the best-rated providers in your city, and does it with far less hassle.

From there, the DogSpotter app makes use of the data it provides to recommend relevant walking services, among them a variety of preferred types of walks for your dog, based on everything from their preferences and sorts of paths to the weather, geographic differences, and degree of safety.

DogSpotter’s founders also contend that their software can — and in some cases already has been — used to make sure that the actual human responsible for dog walking is nearby, by using the location-based data to facilitate long-distance walkings or picking out a dog for a veterinarian who can be contacted.

Speaking of which, they’re not just operating on the basis that it’s an effective way to walk dogs without suffering losses. Although I won’t deny that I’ve seen some big cities suffer from too many small businesses advertising purely online and with little regard for actual human location. As it happens, DogSpotter tells me it has its own network of paying businesses in places like U.S. cities, but it claims to be seeing a company like DogBuddy.com.au as a market that it is disrupting.

My hope is that the company ends up developing a number of other application-specific verticals such as veterinary procedures — again, I have not seen this yet but that would be a great and easy way to monetize.
The business model is to offer DogSpotter services for a set fee per user, per location per month. It includes complimentary services for canine community or adoption groups, dog run assessment and mapping, dog boarding facilities, dog walking packages, and a myriad of other services.

“DogSpotter is different from other dog walking platforms because we have the leading mobile app which is the most effective tool for location-based visual targeting for relevant content and experiences,” said Rowan Zellers, founder and CEO of DogSpotter, in a statement. “This means that DogSpotter is able to provide more relevant products and services and, for early users, we can allow companies to target only the users they want to engage with.”

The problem, of course, is that not all dog walkers have the best apps. DogSpotter says it will compete on performance of its offering, not with other dog walking apps but rather with making it easier for dog owners to find the people they are looking for and, importantly, for those individual dog walkers to save a small bit of money on their operating costs, by saving time and effort with a website. And so on.

}
Generated \headlinefield: DogSpotter raises \$17M to improve your search for walking services

\hrule
\caption{An example of using \modelname-Mega to generate fake business news. This generates an article about a fake startup for `Uber for Dogs', ostensibly created by the first author of this paper.}
\label{fig:uberfordogs}
\end{figure}

\end{document}